\documentclass[a4paper,12pt]{article}

\usepackage{algorithmic}
\usepackage[ruled,vlined]{algorithm2e}

\usepackage{amsmath}
\usepackage{amsfonts}  
\usepackage{amssymb}  
\usepackage[table]{xcolor}
\usepackage{graphicx}

\usepackage{arydshln}
\usepackage{hhline}

\usepackage{geometry}

\title{Learning from scarce information: using synthetic data to classify Roman fine ware pottery}

{\small
\author{Santos J. N\'u\~nez Jare\~no\\
University of Leicester\\
  Leicester, LE1 7RH, UK \\
\and Dani\"el P.  van Helden\\
  University of Leicester\\
  Leicester, LE1 7RH, UK \\
\and Evgeny M. Mirkes\\
  University of Leicester\\
  Leicester, LE1 7RH, UK \\
\and   Ivan Y.~Tyukin\thanks{Corresponding Author:   \texttt{i.tyukin@le.ac.uk};  Tel.: +44-116-252-5106} \\
  University of Leicester\\
  Leicester, LE1 7RH, UK \\
\and Penelope M. Allison \\
  University of Leicester\\
  Leicester, LE1 7RH, UK \\
}
}

\begin{document}

\date{}
\maketitle

\begin{abstract}
In this article we consider a version of the challenging problem of learning from datasets whose size is too limited to allow generalisation beyond the training set. To address the challenge we propose to use a transfer learning approach whereby the model is first trained on a synthetic dataset replicating features of the original objects. In this study the objects were smartphone photographs of near-complete Roman {\it terra sigillata} pottery vessels from the collection of the Museum of London. Taking the replicated features from published profile drawings of pottery forms allowed the integration of expert knowledge into the process through our synthetic data generator. After this first initial training the model was fine-tuned with data from photographs of real vessels. We show, through exhaustive experiments across several popular deep learning architectures, different test priors, and considering the impact of the photograph viewpoint and excessive damage to the vessels, that the proposed hybrid approach enables the creation of classifiers with appropriate generalisation performance.  This performance is significantly better than that of classifiers trained exclusively on the original data which shows the promise of the approach to alleviate the fundamental issue of learning from small datasets.
\end{abstract}

\section{Introduction}

State-of-the-art deep learning models in AI, approaching or even surpassing human's object classifying capabilities (such as \cite{krizhevsky2012imagenet,russakovsky2015imagenet}), require vast training sets comprising millions of training
data \cite{deng2009imagenet, lin2014microsoft}. These requirements are often necessary to  ensure that the models do not merely memorise input-output relationships but also generalise well beyond the data they have been trained on. Indeed, in the simplest binary classification setting with the loss function returning $0$ for correct classification and $1$ otherwise, expected performance $R$ of the classifier, expressed as a mathematical expectation of the loss function, can be estimated as follows \cite{bousquet2003introduction}:
\begin{align}
R\leq R_{\mathrm{train}}+ 2 \sqrt{2 \frac{h \log{\left(\frac{2 e N_{\mathrm{train}}}{h}\right)} + \log{\frac{2}{\delta}}}{N_{\mathrm{train}}}}, \label{eq:risk_estimate}
\end{align}
where $R_\mathrm{train}$ is the empirical mean of the classifier's performance on the training set, $N_{\mathrm{train}}$ is the size of the training set,  $1-\delta$ is the probability that the expected behaviour of the classifier is within this bound, and $h$  is a measure of the classifier's complexity  (Vapnik-Chervonenkis, or VC dimension). 

For a feed-forward network with Rectified Linear Unit (ReLU) neurons, the VC dimension of the whole network is larger than the VC dimension of a single neuron in the network. The latter, in turn, equals the number of its adjustable parameters. Modern deep learning models have thousands of ReLU neurons with  many thousands of adjustable  parameters. For example, ReLU neurons in the fully connected layers of  VGG-19 \cite{simonyan2014very} have $4096$ adjustable weights, and hence the value of $h$ for such networks exceeds $4096$. Thus, if  (\ref{eq:risk_estimate}) is employed to inform our data acquisition processes,  ensuring that the model's {\it expected performance} $R$ does not exceed the value of $R_{\mathrm{train}}+0.1$  with probability at least $0.95$ requires
\[
N_{\mathrm{train}} \geq 35,238,500 >35M
\]
data points. Sharper upper and lower bounds of $h$ for entire networks can be derived from \cite{bartlett1998almost, bartlett2019nearly}. According to  \cite{bartlett1998almost} (Theorem 2.1),  the value of $h$ for a network with $W$ parameters, $k$ ReLU neurons,  and $L$ layers is bounded from above as
\[
h\leq 2 W L \log {\left(2eWL k\right)} + 2 WL^2 \log(2) + 2 L,
\]
which emphasizes the need for large training sets even further.  

Unfortunately, sufficiently large and fully annotated datasets are not available in many applications or research fields. In archaeology, for instance,  datasets almost never reach such orders of  magnitude. Where datasets approaching such a size do exist, they are certainly not comprehensively recorded and photographed, and are not digitally available. Moreover, prior to excavation, these archaeological remains have typically been subjected to various mechanical, chemical, or environmental perturbations that result in their fragmentation (into sherds), introducing a near infinite amount of variability to the dataset that adversely affects statistical properties of any practical/empirical sample. Because of the immense diversity of breaking patterns resulting  from these factors, despite the volumes of pottery collected, even larger datasets are required for the machine to generate the features necessary for classification from these data. Few-shot learning alternatives (methods and networks capable of learning from merely a few examples, such as matching or prototypical networks \cite{vinyals2016matching,snell2017prototypical}), do not address the issue as they either require extensive pre-training on labelled data or assume that data distributions in the AI's latent spaces satisfy some additional constrains \cite{TGAZ2021}, \cite{TGMMT2021} which may or may not hold for a randomly initiated network. Recent successful applications of deep learning in this area \cite{PAWLOWICZ2021105375} exploit identifiable decorative patterns on ceramic vessels to improve ceramic dating processes, but acknowledge the need for larger numbers of typed artefacts for greater accuracy. Also such decorative features are not always available for all ceramics or other types of artefacts.

In this paper we propose and empirically verify through extensive computational experiments that much-needed information for training advanced large-scale AI models, including deep neural networks, can be extracted from abundantly available published knowledge of experts (in our case Roman ceramic specialists) and fed into model training pipelines  \cite{anichiniautomatic}. Not only will this enable us to address the issue of insufficient data but also this will allow us to properly calibrate and control bias in the training data. 

To demonstrate feasibility of the approach we focus on a particular class of artefacts -- pottery remains of a particular fabric type, {\it  terra sigillata}.  {\it Terra sigillata} is a high-quality wheel thrown pottery fabric produced throughout the Roman Empire that can be important for studying cultural differences among past eating and drinking practices \cite{Pim2009}, \cite{Pim2020}. We concentrate on {\it terra sigillata} found in Britain, which was mainly produced in Gaul and the German provinces of the Roman Empire. Focusing on a specific type of artefact enables us to better  illustrate the challenge of bias in real-life training data. This is particularly important because of the fact that among the pottery remains of a particular fabric type (e.g. {\it terra sigillata}) the quantities of each of the available forms are not uniformly distributed across the typological spectrum for that fabric. That is, some forms (or classes) are represented in higher numbers than others. Undoubtedly, this is at least partly the result of past popularity of particular vessel types, and therefore relevant to their diverse uses, or the result of differing physical characteristics of particular classes (some forms are more robust than others), and this might be further complicated by selective collection during post-excavation and then by further biased selection processes in museum collection practices. We are therefore dealing with a dataset that is affected  by various  factors that push the distribution away from the uniform in unpredictable ways.

By basing the perturbations and variations in our simulated models of the artefacts of interest (in our case {\it terra sigillata} vessels) on a widely used British handbook of {\it terra sigillata} \cite{webster1996roman}, we were able to integrate available domain knowledge into our synthetic datasets.  The original objects were photographs which we collected from the Museum of London (MoL) as a part of the  Arch-I-Scan project\footnote{The Arch-I-Scan project, directed by P. Allison and I. Tyukin, out of which the current article flows aims to use thousands of images of pottery fragments to train an AI classifier for Roman fine tablewares focusing on {\it terra sigillata}. This is high-quality wheel thrown pottery fabric produced throughout the Roman Empire. {\it Terra sigillata} found in Britain was mainly produced in Gaul and German provinces of Roman Empire and it is this material the project focuses on.}.  From this collection, being one of the most extensive collections of complete or near complete Roman fine ware vessels in Britain, we took 5373 images of 162 different near-complete {\it terra sigillata} vessels. These original images were used to test and fine-tune our system. Using four popular deep learning architectures, {\it Inception v3} \cite{szegedy2016rethinking}, {\it Mobilenet v2} \cite{sandler2018mobilenetv2}, {\it Resnet50 v2} \cite{he2016deep} and {\it VGG19} \cite{simonyan2014very}, we evaluated our approach by assessing its performance in the task of predicting the vessel class from a single photograph.

The use of synthetic data is very appealing in other areas as well, such as for deep reinforcement learning \cite{sadeghi2016cad2rl}. Multiple synthetic datasets \cite{ros2016synthia, hwang2020eldersim, varol2017learning} or engines  \cite{gupta2016synthetic, deneke2018towards, collins2020traversing} to generate them are now available for AI research. Until recently the domain gap between the synthetic dataset and the real one usually made synthetic-only training non-competitive, but with today’s rendering programs the generalization error is becoming comparable to that between two similar real-life datasets \cite{movshovitz2016useful}. Several procedures (for example generative adversarial networks \cite{bousmalis2017unsupervised, shrivastava2017learning}) have been proposed to deal with the reality gap. In this article, we focus on domain randomization \cite{tobin2017domain, peng2018sim, zakharov2019deceptionnet, mehta2020active} to reduce the gap by encouraging variability of simulation properties, such as object’s colour, illumination, noise, etc. For object detection problems it has been shown that training part of the system on real data and the remaining with realistic synthetic images has a good performance \cite{hinterstoisser2018pre}.

The rest of the article is organised in 4 sections.  Section \ref{Methodology} details data acquisition processes, the  preprocessing of the images, the design of the different experiments and how the different simulated datasets were created. Discussion of the experiment results can be found in Section \ref{results}. Finally, in Section \ref{Conclusions} the conclusions of this work are reported.


\section{Materials and Methods} \label{Methodology}

\begin{figure}[t]
\centering
 \begin{minipage}[b]{0.7\textwidth}
    \includegraphics[width=\textwidth]{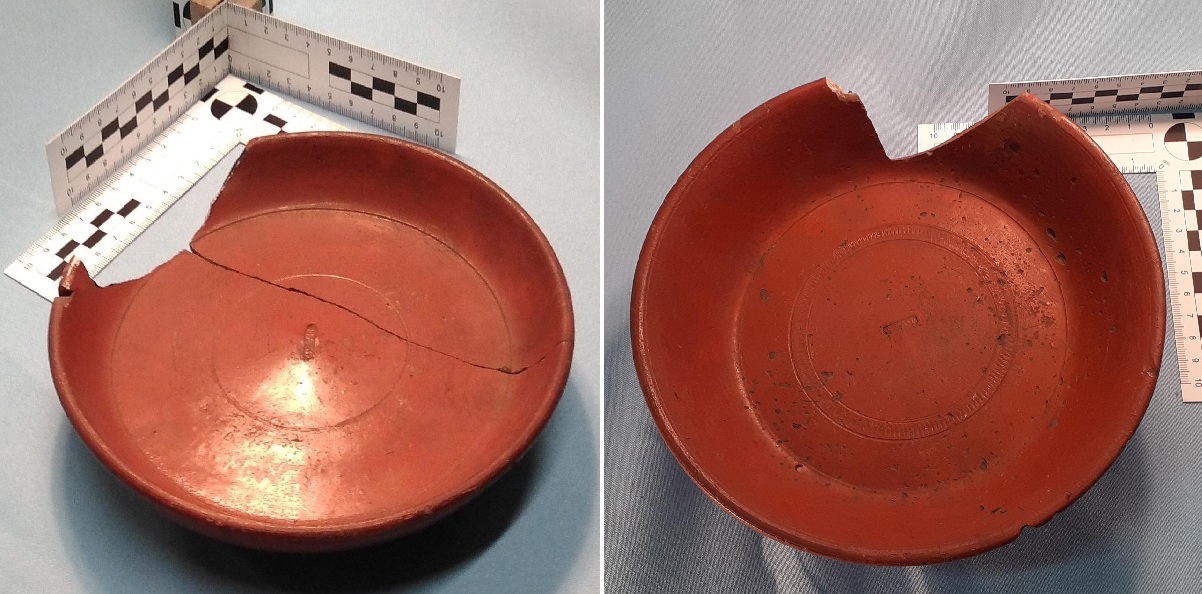}
  \end{minipage}
 \caption{\footnotesize{Example of plain variant of the Dragendorff 18 form (left) and a rouletted variant of Dragendorff 18-31 (18-31R; right). Both are part of our class {\it Dr18}. Photos taken by Arch-I-Scan with permission from the Museum of London. }}
\label{dr18_dr1831R}
\end{figure}

In this section we will explain the procedure used to create and evaluate the performance of an artificial intelligence classifier of the pottery forms in our dataset of images from the MoL collection.  The two main difficulties to overcome were the small number of examples per class and the imbalance in the number of pots in each class in the dataset (see Figure \ref{freq_class}). Even if we produced a lot of images per vessel, we did not have enough different vessels for each class of the dataset to create a classifier for all forms of the established {\it terra sigillata} classification system based on Dragendorff’s work \cite{Dragendorff}. This issue has been alleviated, to some degree, by aggregating some similar forms into one. Furthermore, rouletted (or ``R'') variants of Dragendorff forms have been aggregated into their main form as the main difference is a rouletted impression that is usually only visible in a {\it zenith} view (see Figure \ref{dr18_dr1831R}). For example, our class {\it Dr18} contains Dragendorff forms: 18, 18-31 and 31, as well as their rouletted variants 18R, 18-31R and 31R.

The limited number of pots per class, even where we have combined several Dragendorff forms into a single class, as for {\it Dr18}, limits the possibilities of extensively evaluating architectures and hyperparameter tuning. Thus, we limited ourselves to four standard architectures (see Section \ref{nn_training}) and we evaluated the possibility of improving their performance when pretraining them with different simulated pot datasets. The simulated images of pots were used to augment the dataset of photos we had taken in the MoL collection to make sure the size of the training set was closer to the standards required to train AI classifiers. Notice that the dataset of real pot photgraphs only contains $5373$ images of $162$ different vessels for all pot classes,  whereas the simulation datasets have thousands of images per class with hundreds of different vessels. To ensure the classifier did not become desensitised to the real photos as a result of being overwhelmingly shown synthetic images, it was trained in two phases. In the first, pre-training, phase, the classifier was trained using synthetic images (or left with the initial ImageNet weights in the control condition). After that, a second training was completed using the photos of real vessels. 

In setting up the experiment, the following steps were taken:
\begin{enumerate}
\item The four neural network architectures used in this experiment were modified from their standard and initialised with the ImageNet weights (see Section \ref{nn_training}).

\item Three different sets of simulated pottery vessels were generated so that the impact of (the quality of) simulation on the classifier’s performance could be assessed by comparison. The production of the synthetic datasets is discussed in Section \ref{pot_simulation}.

\item Training, validating, and testing the different neural networks was done using the smartphone photographs of real { \it terra sigillata} vessels from the Museum of London. To make sure these photographs were usable, we created an algorithm which automatically detects the pot, centres it in the photograph and crops out unnecessary surroundings. This process is detailed in Section \ref{pot_detector}.

\item To mitigate the impact of small sample size on our performance metrics, we created 20 different training-validation-test partitions, the creation process of which is detailed in Section \ref{partitions}.

\item  We then trained each of the combinations of four networks and four sets of initial weights with these partitions.

\item The results of each of $16$ combinations of network architectures and pretraining regimes were assessed across the $20$ training-validation-test partitions. The definition of the metrics used for this evaluation is discussed in Section \ref{metrics}, the results themselves are detailed in Section \ref{results}.

\end{enumerate}

\subsection{Data collection} \label{DataCollection}
\begin{figure}[t]
\centering
 \begin{minipage}[t]{0.38\textwidth}
\vspace{0pt}
    \includegraphics[width=\textwidth]{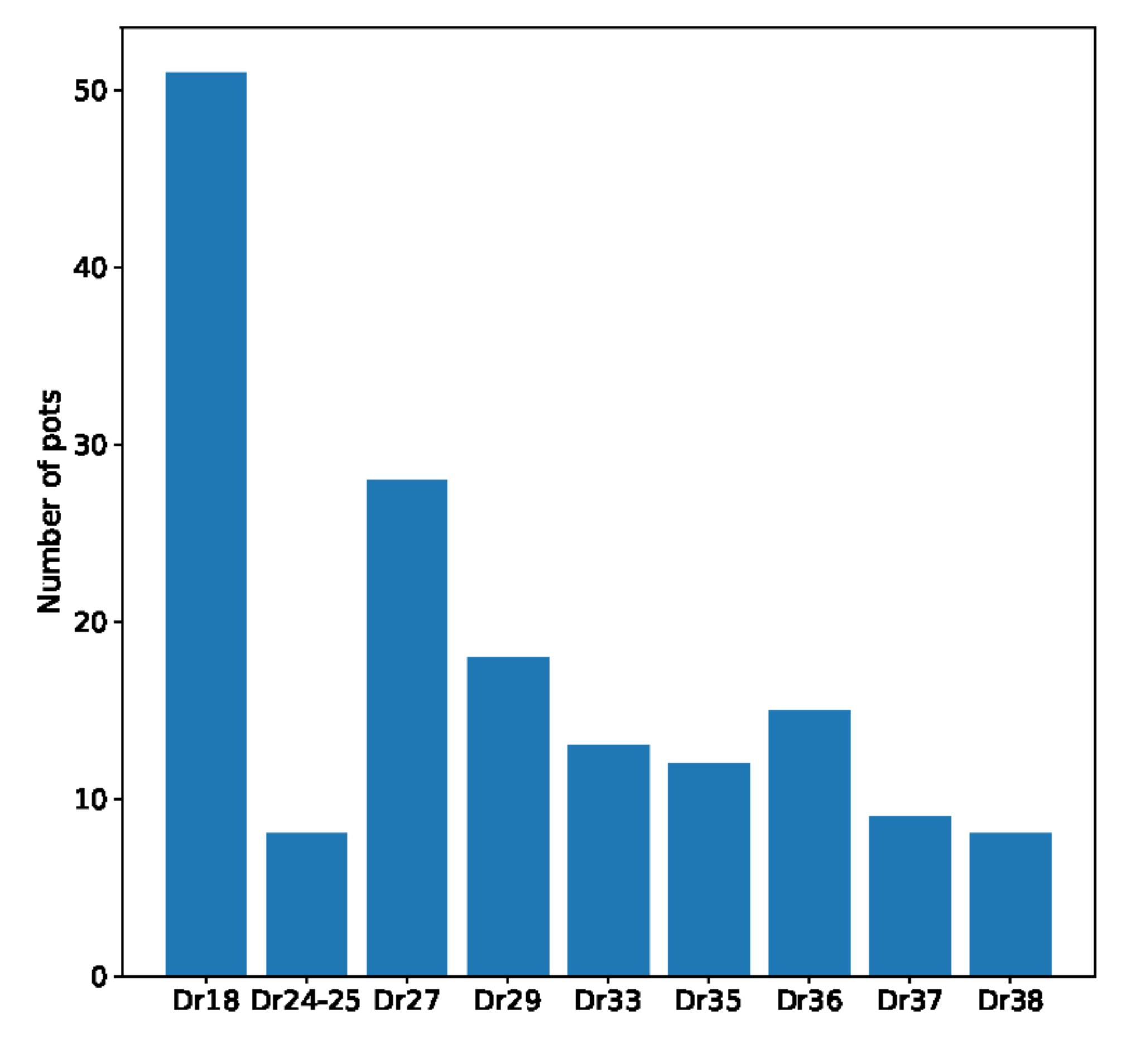}
  \end{minipage}
  \hfill
\begin{minipage}[t]{0.02\textwidth}
\end{minipage}
\hfill
\begin{minipage}[t]{0.2\textwidth}
\centering
\vspace{3pt}
\bgroup
\def\arraystretch{1.43}
\begin{tabular}{|c|c|}
\hline
{\bf Class} &   {\bf \# pots}\\
\hline\hline
 Dr18 &  51 \\
Dr24-25 &   8 \\
    Dr27 &  28 \\
    Dr29 &  18 \\
    Dr33 &  13 \\
    Dr35 &  12 \\
    Dr36 &  15 \\
    Dr37 &   9 \\
    Dr38 &   8 \\
\hline
\end{tabular}
\egroup
\end{minipage}
\hfill
\begin{minipage}[t]{0.05\textwidth}
\end{minipage}

\caption{\footnotesize{Bar Chart and table detailing the number of different pots per class in our dataset. Classes are labelled starting with {\it Dr} followed by the number of the Dragendorff form they are based on  \cite{webster1996roman}. The table gives the exact number of vessels per class.}}
\label{freq_class}
\end{figure}

The vessel photographs were taken by the Arch-I-Scan team, thanks to the project's partnership with the MoL who provided access to its Roman fine ware collection. The whole dataset includes $12395$ images from $384$ different complete or near-complete vessels of different fineware fabrics, and particularly {\it terra sigillata}. For the experiments reported in this article only the {\it terra sigillata} was used ($5373$ images of $162$ vessels). For most of the standard  Dragendorff forms of { \it terra sigillata} (see \cite{webster1996roman} for an introduction to the standard {\it terra sigillata} classification forms in Britain), though, we only had very small numbers of vessels, so these poorly represented forms were also excluded from this experiment. The dataset used here is therefore much smaller than that recorded, being made up of only those {\it terra sigillata} forms for which we had $8$ or more examples  (see Figure \ref{freq_class}).

Photographs were taken within three settings all with the same characteristics except for their position and lighting conditions. That is, each setting comprised a $90$ degree corner wrapped with a light blue cloth to create a chromatic difference from the usual pot colours (i.e. reddish brown for {\it terra sigillata} and dark grey for other London finewares). In order to indicate the sizes of vessels in the photos a set of 3D-scales was used.
Automatic measurement of  vessel size or rim falls outside the scope of this article, however. 

To illuminate each setting, the main light source was provided by overhead reading lamps. In some cases, photographs exhibited a colder illumination when the only light source was the museum basement lights. In order not to have a unique direction of the main light source, the lamp was placed in a different position at each setting.

In each setting photographs were taken by a team of two people. Different smart phones were used throughout the recording session, with different camera resolution and optics.  As required by the MoL regulations, the vessels were carried in and out of  each setting by the museum curator, then one member of the team manipulated each vessel for the different views while the other took the photographs. The following strategy was used to take photos from a range of different views:
\begin{enumerate}
\item With the vessel placed upright on its base and assuming the origin of coordinates is located at the centre of the pot, photographs were taken from azimuth angles of $0$, $45$ and $90$ degrees and  declination angles $0$, $45$ and $90$ degrees. A last photograph with a declination higher than $90$ degrees was taken by resting the mobile on the table.

\item The vessel was rotated an azimuth angle of 90 degrees and the process of point 1 repeated.

\item The vessel was then turned upside down, thus using the rim to support it, and 4 photographs at azimuth $45$ degrees from the declination detailed in point 1 were taken.

\end{enumerate} 
The strategy was generally followed, though deviations from this standard were frequent. This means that while we have standardised photo positions, for certain vessels we have many more photos than described in the process above and for a few vessels we have fewer. This introduces some imbalances in the dataset.

\begin{figure}[!tbp]
\centering
 \begin{minipage}[b]{0.9\textwidth}
    \includegraphics[width=\textwidth]{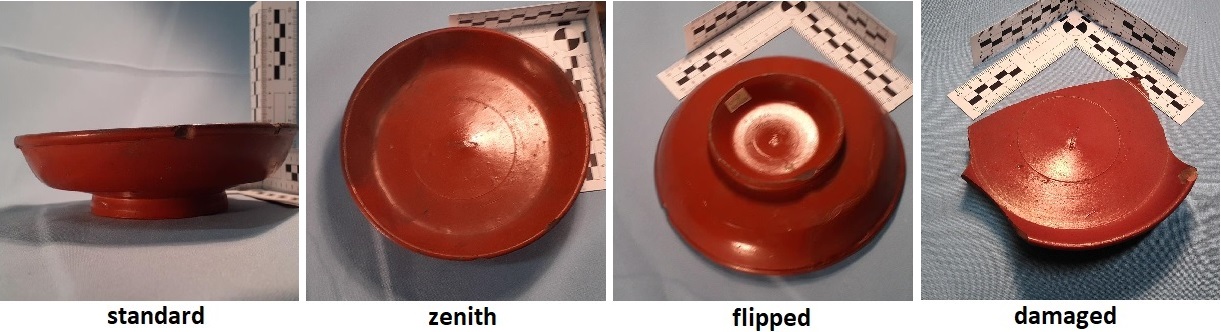}
  \end{minipage}
 \caption{\footnotesize{Example of the three different perspectives ({\it standard, zenith, flipped}; seen in the three left-most images) and the {\it damaged} condition (right-most image) using photos from class {\it Dr18}. Note that images have already been through the automatic cropping procedure detailed in Section \ref{pot_detector}. Photos taken by Arch-I-Scan with permission from the Museum of London.}}
\label{types_photos}
\end{figure}

As a final step, images were manually labelled (see Figure  \ref{types_photos}). As well as the museum inventory number for the pot this label includes information about the perspective from which the photo was taken or the vessel condition. The {\it standard} label was given to those vessels that were standing in their natural orientation and photographed from declination angles of 45º, 90º and greater than 90º. This includes, but is not restricted to, the profile perspective generally used in archaeology. The {\it zenith} label was given to those photographs that were taken from above with the vessel standing upright as normally. The goal was to achieve angle 0º, but as can be seen from Figure  \ref{types_photos}, this angle was not always perfect. All those photographs in which the vessel was supported by its rim, irrespective of the angle of the photograph, were given the label {\it flipped}. Finally, the label {\it damaged} was given to those vessels with half their azimuth or more in a poor condition.

\subsection{Neural nets configurations}\label{nn_training}

As we have a very limited number of pots per class, a robust exploration of different architectures and finetuning their hyperparameters is not possible. Therefore, we limited ourselves to standard architectures used in image classification problems, namely: {\it Inception v3} \cite{szegedy2016rethinking}, {\it Mobilenet v2} \cite{sandler2018mobilenetv2}, {\it Resnet50} \cite{he2016deep} and {\it VGG19} \cite{simonyan2014very}. We used the {\it ImageNet} problem \cite{deng2009imagenet, russakovsky2015imagenet, krizhevsky2012imagenet} initialization weights, though we pretrained with three different simulated datasets in order to improve the performance through domain adaptation.  We will show how the pretraining with synthetic pot images results in improvement in the accuracy of the model that increases as we consider more realistically simulated pots (the details of the simulated datsets are discussed in Section \ref{pot_simulation}).

\begin{figure}[t]
\centering
 \begin{minipage}[b]{0.5\textwidth}
    \includegraphics[width=\textwidth]{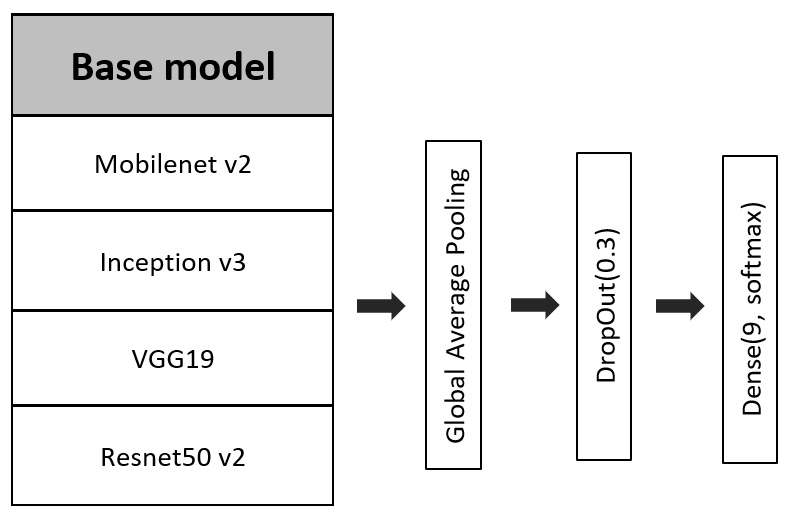}
  \end{minipage}
 \caption{\footnotesize{Diagram of the neural nets considered in the article. Four different base architectures are considered: {\it Mobilenet v2, Inception v3,  Resnet50 v2} and {\it VGG19}. The last layers of these architectures were removed up to the convolutional structure and substituted by a Global Average Pooling layer followed by a Drop Out layer with $0.3$ drop probability and a final dense layer with softmax activation and $9$ nodes (one for each of the 9 pot classes). }}
\label{nn_config}
\end{figure}

 With the exception of the image size ($224\times224$ for {\it Mobilenet v2}, {\it Resnet50 v2} and {\it VGG19};  and $299\times299$ for {\it Inception v3}), all nets have been trained with the same configuration (see also Figure \ref{nn_config}): 
\begin{itemize}
\item  Backbone convolutional neural net base architecture (i.e. the last layers of these architectures were removed until the convolutional structure)
\item A global average pooling layer after the convolutional structure.
\item A drop out layer with 0.3 exclusion probability 
\item A final bottleneck dense layer with softmax activation, i.e a linear dense layer followed by a softmax transformation of the outputs.
\end{itemize}


\subsection{Pot simulations}
\label{pot_simulation}

To pretrain the neural nets into a domain closer to our problem, we developed two procedures to generate datasets of simulated pots. The first one uses the Python package Matplotlib \cite{hunter2007matplotlib} to generate a simple form of the different classes. The other uses Blender \cite{blender}, an open source 3D modelling tool, to easily create simulated images very close to real images.

\begin{figure}[t]
\centering
 \begin{minipage}[b]{\textwidth}
    \includegraphics[width=\textwidth]{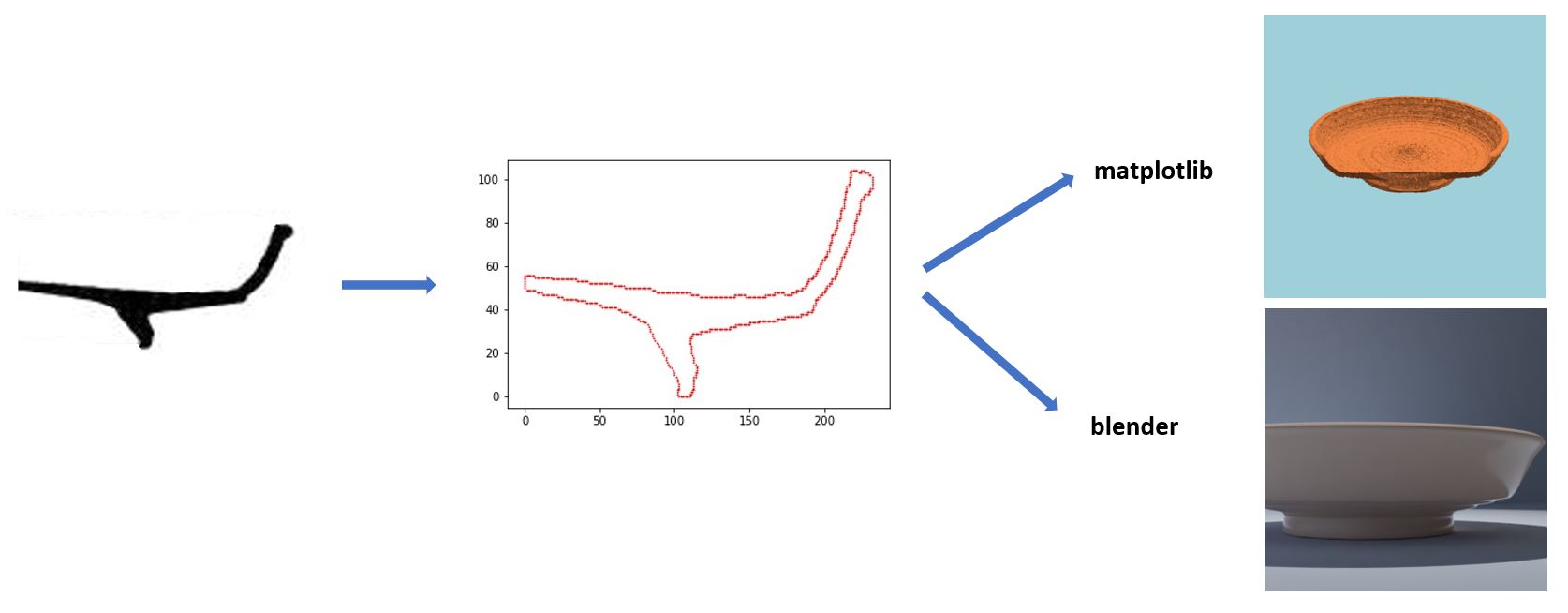}
  \end{minipage}
 \caption{\footnotesize{Diagram of the pot simulation. Starting from a digitized profile drawing the profile is extracted as an ordered list of points. With that profile we generated simulated photographs  using  the package {\it Matplotlib} or the 3D modelling tool {\it Blender}. }}
\label{pot_sim}
\end{figure}

Both procedures take as input the profile of each vessel. In order to obtain these profiles we scanned Webster's \cite{webster1996roman} profile drawings corresponding to the Dragendorff forms present in our dataset. We manually erased all the details that were not part of the rotational symmetric shape and thereby created a collection of digitised black and white drawings of these pot profiles. So, if we define $x_{i,j} \in \{0,1\}$ as the pixel in position $(i,j)$ with only two possible colors/values: $0$ for white/background and $1$ black/profile, the border pixel set can be detected as:
\[
\text{border} \equiv \{ (i, j):  \vert x_{i,j} - x_{i-1,j}\vert + \vert x_{i,j} - x_{i,j-1}\vert  >0 \},
\]
which basically computes the vertical and horizontal differences and sums their absolute values. A single difference would not be enough as wherever the border follows a vertical or horizontal trajectory some pixels would not be detected. Finally, we merely have to order the border set, resulting in a list in which for each element its two neighbours also correspond to the two pixels closest to its position $(i,j)$. The list of points is used by each procedure to draw an axial section, which is rotated to generate a 3D mesh of points that corresponds to the rotationally symmetric shape of each vessel.  Figure \ref{pot_sim} shows a diagram of this process (see \cite{itkin2019computational}, \cite{anichini2020developing}, \cite{anichiniautomatic} for a comparable process).

\begin{figure}[t]
\centering
 \begin{minipage}[b]{0.7\textwidth}
    \includegraphics[width=\textwidth]{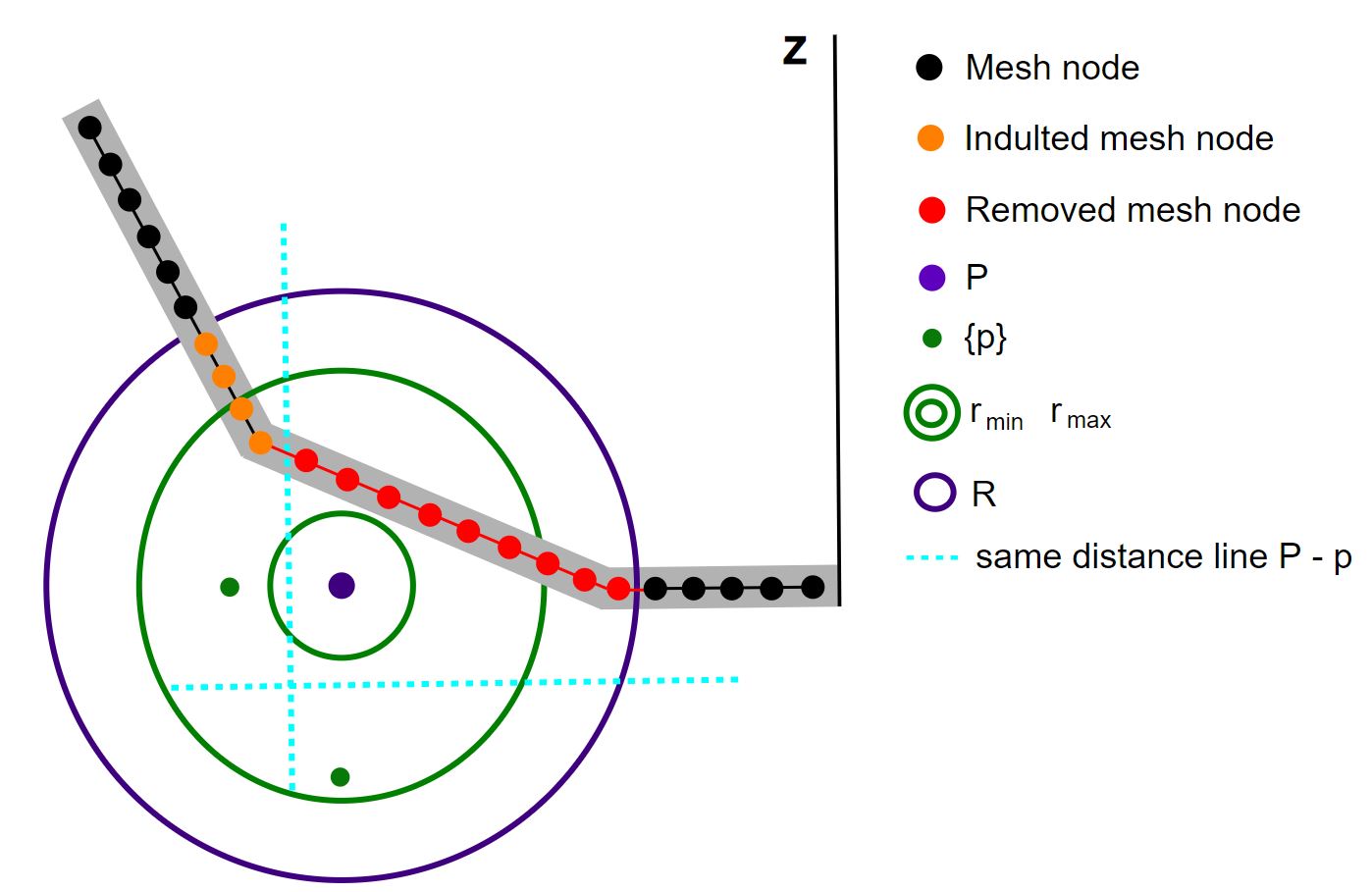}
  \end{minipage}
 \caption{\footnotesize{2D diagram of the breaking procedure. Mesh points that fall within radius R (purple circle) from P (purple point) are susceptible to being removed. Points {p} (green points) are inside the crown of radii $r_{\text{min}}$ and $r_{\text{max}}$ (green circles), the lines that separate mesh points closer to $\{p\}$ than $P$ are indicated as a dashed cyan line. Black points in the pot mesh are not susceptible to be removed; orange ones are susceptible, but have been pardoned because they are closer to a green point $\{p\}$ than to $P$ (purple); and lastly, red ones have been removed. }}
\label{breaking_scheme}
\end{figure}

As we have previously discussed, almost all photographed real pots have at least some degree of damage. In order to simulate such breaks we designed a similar automatic breaking procedure for both software programs. Each break was generated by choosing a random position, $P$,  near the pot mesh.  The points of the mesh inside a sphere of radius $R$ from $P$ were susceptible to being removed. After that, we randomly chose $n$ points, $\{p\}$, between distances $r_{\text{min}}$ and $r_{\text{max}}$ from $P$. We then erased any part of the vessel which is inside the sphere of radius $R$ and closer  to $P$ than to a new point, $\{p\}$. This procedure is implemented in different ways depending on the software used. While in Blender the points are actually removed from the mesh using the Boolean transformation tool, when using Matplotlib we set  the alpha (opacity) channel of these points facets to zero. By modifying the number of breaks and the different parameters ($R$, $n$, $r_{\text{min}}$ and $r_{\text{max}}$) we created different random breaking patterns that gave us enough variability to create a good simulated training dataset. Figure \ref{breaking_scheme} shows a diagram of this process.

\begin{figure}[hbtp]
\begin{minipage}[b]{0.9\textwidth}
\centering
 \begin{minipage}[b]{0.7\textwidth}
    \includegraphics[width=\textwidth]{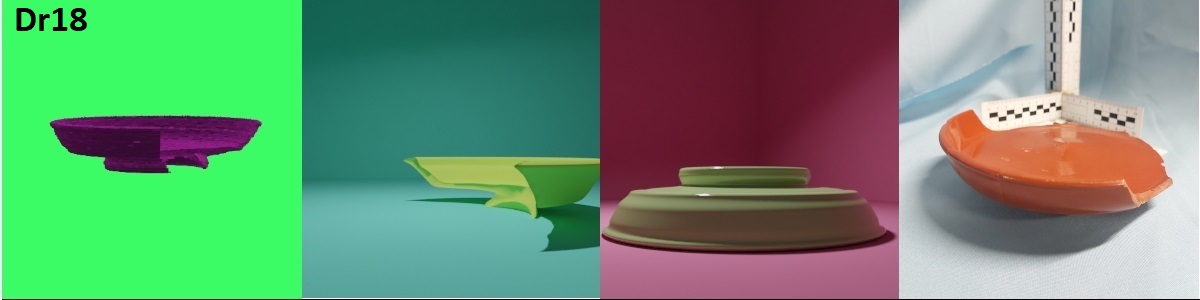}
  \end{minipage}
\begin{minipage}[b]{0.7\textwidth}
    \includegraphics[width=\textwidth]{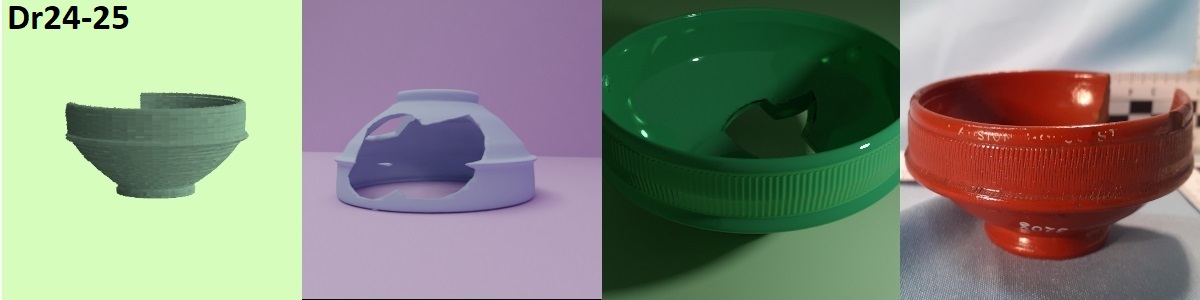}
  \end{minipage}
\begin{minipage}[b]{0.7\textwidth}
    \includegraphics[width=\textwidth]{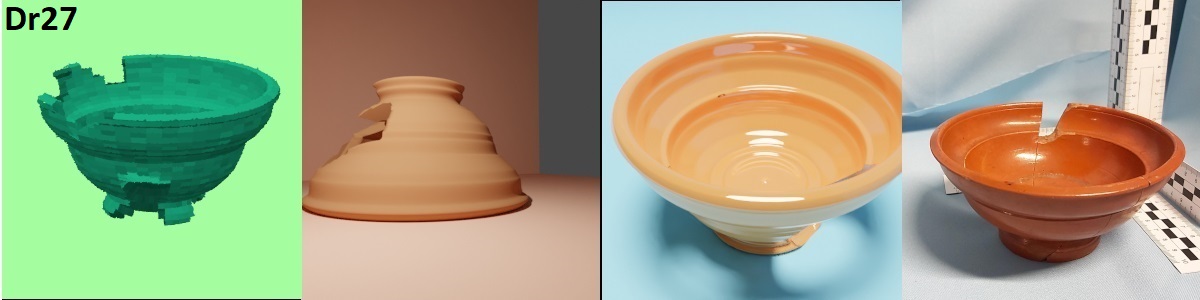}
  \end{minipage}
\begin{minipage}[b]{0.7\textwidth}
    \includegraphics[width=\textwidth]{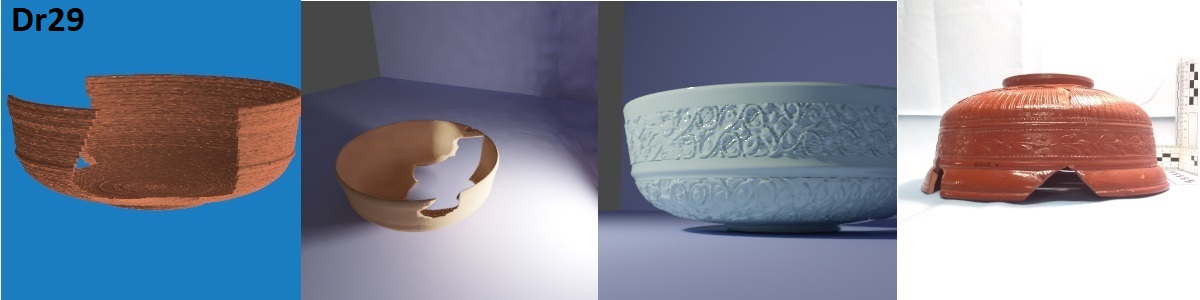}
  \end{minipage}
\begin{minipage}[b]{0.7\textwidth}
    \includegraphics[width=\textwidth]{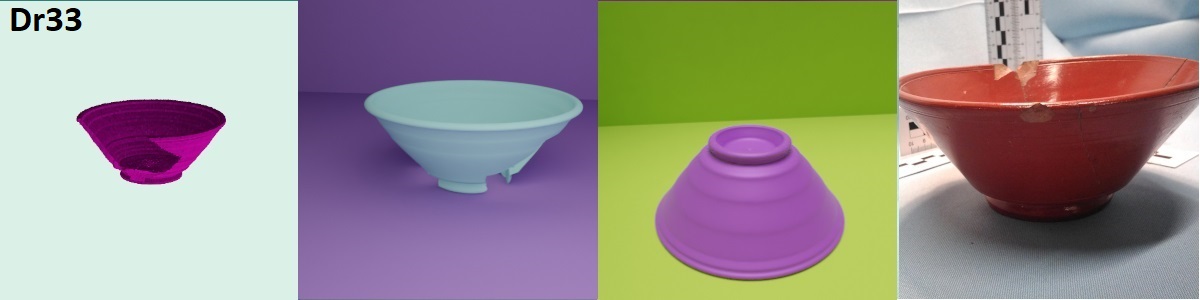}
  \end{minipage}
\begin{minipage}[b]{0.7\textwidth}
    \includegraphics[width=\textwidth]{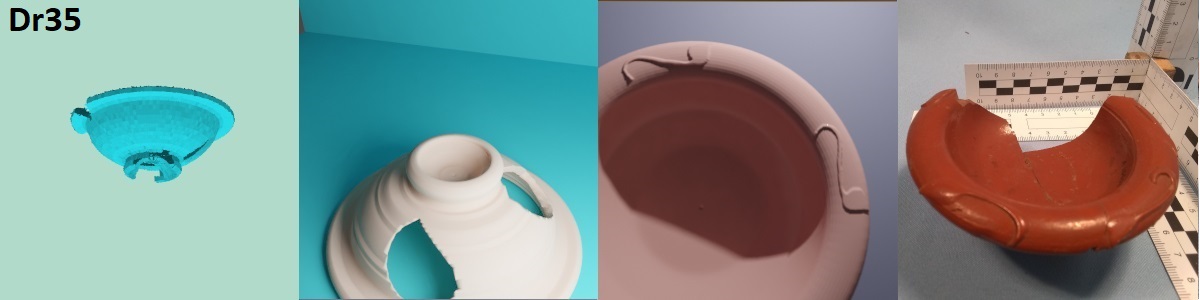}
  \end{minipage}
\begin{minipage}[b]{0.7\textwidth}
    \includegraphics[width=\textwidth]{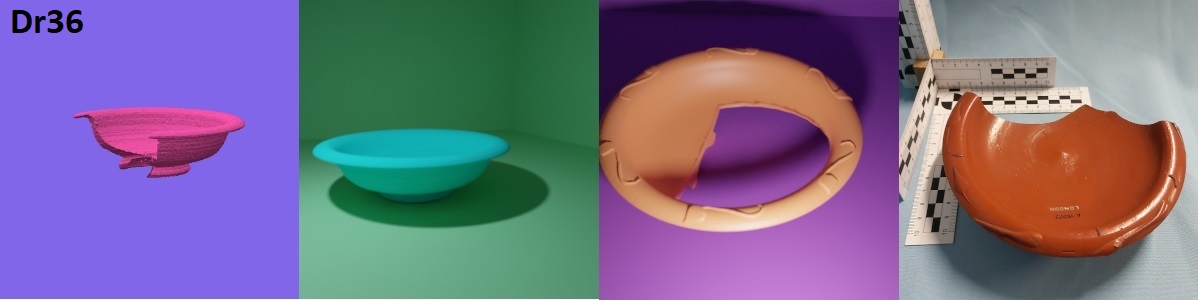}
  \end{minipage}
\begin{minipage}[b]{0.7\textwidth}
    \includegraphics[width=\textwidth]{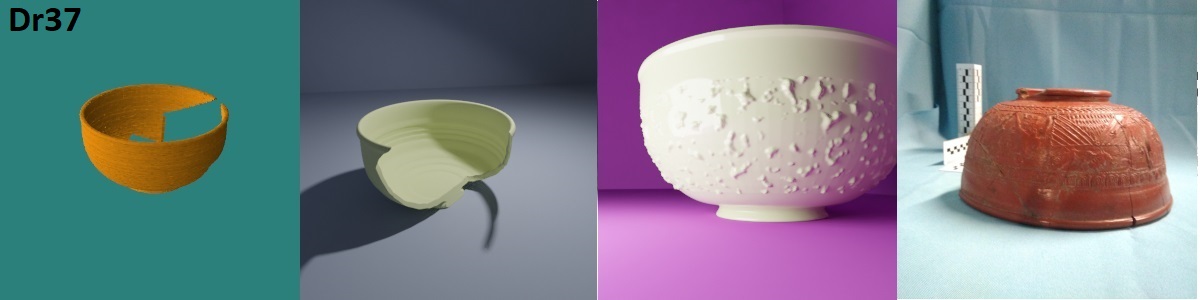}
  \end{minipage}
\begin{minipage}[b]{0.7\textwidth}
    \includegraphics[width=\textwidth]{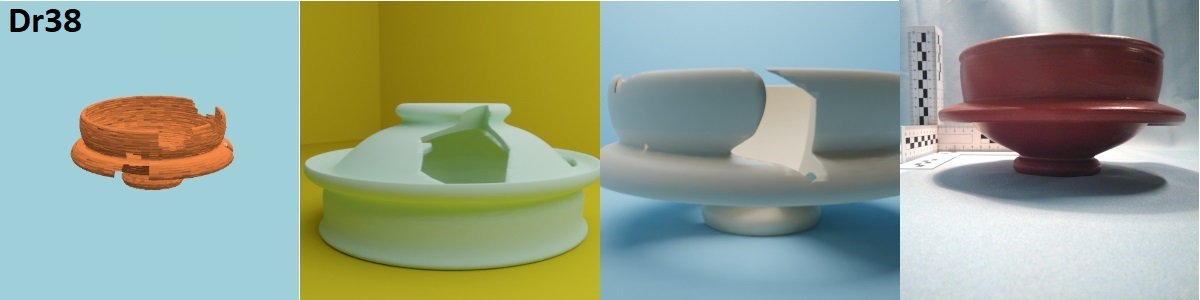}
  \end{minipage}
\end{minipage}
\caption{\footnotesize{Examples for all simulated datasets and real photograph of our $9$ different {\it Dr} classes. From left to right: matplotlib, blender1, blender2, real photo. Photos taken by Arch-I-Scan with permission from the Museum of London.}}
\label{classes_imgs}
\end{figure}

We have created three different simulated datasets:
\begin{itemize}
\item {\bf matplotlib} ($1000$ images per class): 
\newline
To generate this dataset the Python package Matplotlib \cite{hunter2007matplotlib} was used. The simulated pot was floated in a homogeneous background, no shadow is projected but the pot colour is affected by the light. The light source is a point far away so only the angle has been changed. The profiles were softened to avoid their small defects creating circular patterns that could be identified by the neural net. For the same reason, a small amount of random noise was added to the surface mesh points position. 

\item {\bf blender1} ($1400$ images per class): 
\newline
This  dataset was generated using Blender \cite{blender}. We built a background set close to the original photography setting with a uniform colour. The lighting conditions were randomly chosen using the different lighting classes available, namely: sun, spot, point. Thanks to the rendering feature {\it Cycles}, we could  simulate shadow projections as well as changes of illumination in both pot and setting. The profile was softened by distance using the software and no noise was added. The material properties where not changed from the default ones except for the colour. 

\item {\bf blender2} ($1300$ images per class): 
\newline
The process followed to create this dataset is similar to the one used to create {\it blender1}, however, the material properties were changed to make them more similar to {\it terra sigillata} pots. Thus,  some images  exhibit  simulated pots with a reflective material which creates light-saturated regions in the image of the pot. We have also added decoration and motifs to some classes, namely:
\begin{itemize}
\item { \it Dr24-25}: half of the images show a striped pattern near the top.
\item { \it Dr35} \&{\it Dr36}: half of the images show a pattern of four or six leaves on their rim.
\item { \it Dr29} \&{\it Dr37}: The decoration has a lot of variability in reality. We have simulated it in half of the images through two noise pattern displacements (Blender {\it Musgrave} and {\it Magic} textures).
\end{itemize}
Finally, we sieved each synthetic dataset, removing images that were taken from too close or show some defects resulting from the simulated breaking procedure.
\end{itemize}
Some samples of the different simulation datasets for all  the Dragendorff pot forms included in this study can be seen in Figure \ref{classes_imgs}.

\subsection{Pot detector: cropping and centering the images}
\label{pot_detector}

Although not the main focus of this article, cropping and centering the pot image is a vital step in order to have a good automatic classifier. Object detection is a field that has seen huge development in recent years \cite{zhao2019object}. As with image classification problems these developments have been propelled by convolutional neural networks. Several architectures and approaches can be considered for an object detection problem, but they should basically cover three steps \cite{zhao2019object}: generating crops of the image, extracting features out of each crop, and assessing which class the object belongs to. 

The controlled conditions in which our photographs were taken mean that our task is relatively simple compared to the complexity of problems which state-of-the-art object detection algorithms were designed to tackle. Due to the uniformity of the image background and the homogeneity of colour, {\it terra sigillata} vessels can be located using colour histograms. As we have deliberately not standardised the lighting conditions, however, some parts of the pot are in shadow. To deal with some of the errors introduced in some photographs by these factors, we used a different method. It is still based on colour properties and position rather than on pot-shape learning, though.

The  algorithm first reduces the image size to $30\times30$ pixels, this particular size was manually chosen. For each pixel we transform its colour to the Hue-Saturation-Value basis. This kind of pottery has a high Saturation and Value in comparison with the background.  For each image a dataset was created where each row corresponds to a pixel and includes the pixel position in the x-dimension ($\text{pos}_x$) and y-dimension ($\text{pos}_y$), to encourage spatially compact groups. Thus, the final features are:  Saturation, Value, $\text{pos}_x$ and $\text{pos}_y$.  The features are linearly normalized into the $[0,1]$ interval. Using the DBSCAN algorithm \cite{ester1996density, schubert2017dbscan} with the maximum distance between two pixels to be considered in the same neighbourghood set to $\text{Eps} = 0.1$ and the number of points in the neighboughood (including itself) to be considered a core point set to $\text{MinPts}=5$, we can group the pixels of the image. After that, we filter those groups with a small number of pixels, a threshold of $25$ pixels was manually tuned. Finally, we exploit the fact that the pot is usually centered in the photograph and compute for each group the average distance to the center of the photograph. We select as the pot group that with the minimum average distance. 

Once the pot has been located we make a square crop in the original image with the pot cluster's mean position, $(\overline{pos}_x, \overline{pos}_y)$, in its center. Note  that the location of each pixel was scaled to the image original size. Also note that in order to do that we calculate the smallest rectangle containing the cluster and get its larger side, $L_0$. We increase the initial side length, $L = 1.5\cdot L_0$, to avoid cropping the borders of the pot. An outline of the procedure to locate and crop the pot can be found in Algorithm \ref{algorithm_pot_detection}.

\begin{algorithm}[t]
\SetAlgoLined
\hspace*{\algorithmicindent} \textbf{Input: } {Image} \\
\hspace*{\algorithmicindent} \textbf{Output: } {Image\_output. Cropped and centered pot image.}
\vspace{3mm}

 scaled\_image = {\bf  scale} input Image to $30 \times 30$\;
scaled\_image = {\bf change color basis} from RGB to HSV\;
dataset\_pixels = {\bf create dataset} where each pixel (rows)  in scaled\_image have features: [Saturation, Value, $\text{position}_x$, $\text{position}_y$]\;
dataset\_pixels = {\bf linearly normalize} dataset\_pixels features into [0,1] interval\;
pixel\_group = {\bf assign cluster} to each pixel using DBSCAN(Eps=0.1, MinPts=5) algorithm\;
dataset\_pixels = dataset\_pixels {\bf remove} pixels of groups with less than 25 pixels\;
group\_distance = for all groups compute {\bf mean}($(\text{position}_x - 0.5)^2 + (\text{position}_y - 0.5)^2$ )\; 
pot\_group = {\bf argmin}(group\_distance)\;
dataset\_pixels = dataset\_pixels {\bf remove} pixels not in pot\_group\;
$L_x$ = {\bf  max}($\text{position}_x$)) - {\bf min}($\text{position}_x$))\;
$L_y$ = {\bf max}($\text{position}_y$)) - {\bf min}($\text{position}_y$))\;
{\bf scale} $L_x$  and $L_y$ to the original size\; 
L = $1.5\; \cdot$ {\bf max}($L_x$, $L_y$)\;
L\_half = {\bf integer\_part}(L/2)\;
$\text{center}_x$ = {\bf mean}($\text{position}_x$)\;
$\text{center}_y$ = {\bf mean}($\text{position}_y$)\;

Image\_output = Image[{\bf from} $\text{center}_x$  - L\_half {\bf to} $\text{center}_x$  + L\_half,
{\bf from} $\text{center}_y$  - L\_half {\bf to} $\text{center}_y$  + L\_half]\;

\caption{Locate and crop pot images}\label{algorithm_pot_detection}
\end{algorithm}

\subsection{Training-validation-test partitions}
\label{partitions}

The limited number of vessels we have means that a classifier's performance can be very dependent on whether a certain vessel is allocated to the training, test, or validation set. This leads to a high variance in the performance metrics and, as a result, low confidence in the reliability of estimations of the classifier's accuracy. To ensure a more robust estimation of the fitted model’s accuracy and variance, we have generated 20 different training-validation-test partitions and consider the model's performance across these partitions. For each training set we have randomly chosen four different vessels per class. Another two vessels per class were used for validation, and the remaining vessels were used as the test set. By having the same number of vessels for each class we reduce the risk of an unbalanced training, though the number of different vessels available for training is small. We also have a larger test set and, therefore, more reliable error estimations. We used the same 20 training-validation-test partitions to train each of the four network architectures. The average performance across the partitions was used to evaluate the effect of the quality of simulated vessels (see Section \ref{pot_simulation}) on the overall performance of the model.

In order to train the architectures, we have chosen to minimize the the categorical cross entropy loss training all the layers. As was decided during an initial exploration, we used an Adam \cite{kingma2014adam} optimizer with learning rate $5\cdot10^{-5}$ (except for {\it VGG19} that is reduced to $5\cdot 10^{-6}$). The batch size was chosen to be $8$ to fit in our computational resources and a patience of 10 epochs was set to stop the training. The weights of the epoch with best validation categorical cross entropy were selected.

\subsection{Performance metrics}
\label{metrics}

The objective of this work is to create an algorithm, $\widehat{G}^\mathcal{D}: \mathcal{X} \rightarrow \mathcal{G}$ , trained with the training dataset $\mathcal{D}$, that, given an image $x$ in the space of pot images, $x \in \mathcal{X}$, maps it into $g \in \mathcal{G}$, where $\mathcal{G}$ is the set of pot  classes. The training dataset is a random sample from some probability distribution $\mathcal{D}  \equiv \{(x, v, g)\} \sim P_{\text{train}}(X, V, G)$ such that $X$ is random variable representing the image, $V$ is a random variable that takes values in $\mathcal{V}$ the set of all pots (vessels) and $G$ represents the class it belongs to.

Let $f^\mathcal{D}: \mathcal{X} \rightarrow  \mathbb{R}^{\text{card}(\mathcal{G})}$ represent the neural net trained with the training dataset $\mathcal{D}$ that takes an image as input and returns a vector with the prediction scores for each class, $f^\mathcal{D}_i(X);\;\; i \in\{1,\ldots,\text{card}(\mathcal{G})\}$ (the cardinality, $\text{card}$, is the number of elements of a set; in our case $\text{card}(\mathcal{G}) = 9$, the number of pot type forms in our dataset), thus we can define our classifier as,
\[
\widehat{G}^\mathcal{D}(X) =  \arg \max_i f^\mathcal{D}_i(X).
\]
The performance of the algorithm can be assessed using the test dataset $\mathcal{D}^{t}\equiv \{(x, v, g)\} \sim P_{\text{test}}(X, V, G)$. With $\delta_{i,j}$  being the Kronecker delta (which is $1$ if $i=j$ and $0$ otherwise) and $E$ the expected value, we can define the accuracy of $\widehat{G}^\mathcal{D}$ in $P_{\text{test}}$ as
\begin{align}
\text{acc}[\widehat{G}^\mathcal{D}, P_{\text{test}} ] \equiv E_{(x,v,g) \sim P_{\text{test}} }\left[ \delta_{g,\;\widehat{G}^\mathcal{D}(x)}\right].
\label{acc_test}
\end{align}
Given a test dataset, $\mathcal{D}^t$, we can write an estimate for this accuracy as,
\[
\widehat{\text{acc}}\left[\widehat{G}^\mathcal{D},\mathcal{D}^t \right] = \frac{1}{\text{card}(\mathcal{D}^t)} \sum_{(x,g) \in \mathcal{D}^t } \delta_{g,\;\widehat{G}^\mathcal{D}(x)}.
\]

We would like to estimate accuracy (\ref{acc_test}) for the hypothetical distribution, $P_{\text{test}}(X, V, G)$, of a final user testing the classifier on pots not previously seen by the classifier. However, we only  have the dataset that we have collected, which is imbalanced in both number of different pots per class (see Figure \ref{freq_class}) and number of photos per pot (from a minimum of 14 to a maximum of 100). If we simply consider each photo as a random sample our metric will depend on the imbalances mentioned. Thus we have to make assumptions about the real distribution $P_{\text{test}}(X, V, G)$ and modify the sampling of our test dataset accordingly. Firstly, we decided that the test sampling probability must not privilege any particular pot or photograph. $\text{Unif}(Z,\mathcal{Z})$ being the uniform distribution of $Z$ over the set $\mathcal{Z}$ that means, 
\begin{align}
P_{\text{test}}(X,G, V) =& \;P_{\text{test}}(X \vert G, V) \cdot P_{\text{test}}(V\vert G) \cdot  P_{\text{test}}(G) \nonumber\\
=& \; \text{Unif}\left(X, \left\{ x \in \mathcal{X} \vert (x,v,g) \in \mathcal{D}^t; v=V; g=G\right\}\right) \nonumber \\
&\cdot \text{Unif}\left(V, \left\{ v \in \mathcal{V} \vert (x,v,g) \in \mathcal{D}^t; g=G\right\}\right) \cdot P_{\text{test}}(G).
\label{assumption1}
\end{align}
Secondly, we have assigned $P_{test}(G)$ considering two scenarios:
\begin{itemize}
\item {\bf Uniform prior:} all classes have the same probability and, thus, are equally weighted in the final metrics.
\begin{align}
P_{\text{test}}(G) = \frac{1}{\text{card}(\mathcal{G})} = \frac{1}{9}. 
\label{uniform_prior}
\end{align}

\item {\bf MoL prior: } We assume that the MoL pot classes distribution is a good estimator of  $P_{\text{test}}(G)$. Let $\mathcal{D}^{\text{dataset}}$ be the full dataset of our collection, the probability will be given by the number of pots of class $i$ over the total number of pots:
\begin{align}
 P_{\text{test}}(G=i) = \frac{ \text{card} ( \{v \in \mathcal{V}\vert \exists (x,v,g) \in \mathcal{D}^{\text{dataset}};  g=i \})}{ \text{card} ( \{v \in \mathcal{V}\vert \exists (x,v,g) \in \mathcal{D}^{\text{dataset}} \})}. 
\label{museum_prior}
\end{align}
\end{itemize}
Each prior provides us with an interesting perspective. On the one hand, the uniform prior assesses how well the algorithm has learnt the classes without prioritising any one class. On the other hand,  were the MoL prior close to the field frequency of the classes, it would give us scores closer to the user performance perception who would find more frequent classes more often.

Thus, given a test dataset of our collection, $\mathcal{D}^{\text{test-collection}} \equiv \mathcal{D}^{\text{t-c}} $, we will modify the sampling probabilities to fulfill either (\ref{assumption1}) and (\ref{uniform_prior}), or (\ref{assumption1}) and (\ref{museum_prior}) . If $n^{(\text{photos})}_{i j} = \text{card}(\{(x,v,g) \in \mathcal{D}^{\text{t-c}}\vert g=i;v=j\})$ is the number of photos of pot $j$ that belongs to class $i$ and $n^{(\text{pots})}_i = \text{card} ( \{v \in \mathcal{V}\vert \exists (x,v,g) \in \mathcal{D}^{\text{t-c}};  g=i  \})$ is the number of pots of class $i$, we can write the sampling probability for the test dataset of our collection as,
\begin{align}
P_{\text{samp}}&(X=x, V=v, G=g) = P_{\text{samp}}(X=x \vert V=v, G=g) \nonumber\\
& \cdot  P_{\text{samp}}(V=v \vert G=g) \cdot  P_{\text{samp}}(G=g) \nonumber\\
&= \frac{1}{n^{(\text{photos})}_{g v}} \cdot \frac{1}{n^{(\text{pots})}_g }\cdot  P_{\text{test}}(G=g);\;\;\; \forall (x,v,g) \in \mathcal{D}^{\text{t-c}}, \nonumber
\end{align}
where $ P_{\text{test}}(G)$ follows one of the prior distributions assumed, namely the uniform prior (\ref{uniform_prior}) or the MoL prior (\ref{museum_prior}).

This way we can obtain an expression for the estimated accuracy:
 \begin{align}
\widehat{\text{acc}}\left[\widehat{G}^\mathcal{D},  P_{\text{samp}} \right] = \sum_{(x,v,g) \in \mathcal{D}^{\text{t-c}} }  \frac{1}{n^{(\text{photos})}_{g v}} \cdot \frac{1}{n^{(\text{pots})}_g }\cdot  P_{\text{test}}(G=g) \cdot \delta_{g,\;\widehat{G}^\mathcal{D}(x)}. \label{acc_single}
\end{align}

One of the main results of this paper is evaluating the effect of pretraining with synthetic datasets on the performance of the classifier. However, our very limited number of pots would impede reliable estimations of the effect if we restrict ourselves to a single test set. To obtain a more robust estimation let us define an equivalence relation, $\mathcal{R}$,  between training datasets randomly generated through the procedure described  in Section \ref{partitions}. Let the set of all equivalent training datasets be represented by $\mathcal{D}/\mathcal{R}$, using (\ref{acc_test}) the expected accuracy for an equivalent training dataset can be defined as,
\begin{align}
\overline{\text{acc}}\left[\widehat{G}, P_{\text{test}} \right] \equiv E_{\mathcal{D}/\mathcal{R}}\left[ \text{acc}[\widehat{G}^\mathcal{D}, P_{\text{test}} ]\right] = E_{\mathcal{D}/\mathcal{R}}\left[E_{(x,g) \sim P_{\text{test}} }\left[ \delta_{g,\;\widehat{G}^\mathcal{D}(x)}\right]\right]. \label{overlineacc}
\end{align}
By generating $n_{\text{splits}} = 20$ train-validation-test splits as explained in Section \ref{partitions}, we can give an estimation of (\ref{overlineacc}). $\mathcal{D}_i$ being the training set  and $\mathcal{D}^{\text{t-c}}_i$ the test dataset for each partition $i \in \{1,\ldots,n_{\text{splits}}\}$, the estimation can be written as
\begin{align}
\widehat{\overline{\text{acc}}}\left[\widehat{G}, P_{\text{samp}} \right] =\; &\frac{1}{n_{\text{splits}}} \sum_{i=1}^{n_{\text{splits}}}\sum_{(x,v,g) \in \mathcal{D}^{\text{t-c}}_i }  \frac{1}{n^{(\text{photos})}_{g v}} \cdot \frac{1}{n^{(\text{pots})}_g }\nonumber \\
&\cdot  P_{\text{test}}(G=g) \cdot \delta_{g,\;\widehat{G}^{\mathcal{D}_i}(x)}. \nonumber
\end{align}
This quantity estimates the bias of the model $\widehat{G}$ when trained with a dataset in $\mathcal{D}/\mathcal{R}$. The variance,
\[
\sigma^2_{\text{acc}\left[\widehat{G}, P_{\text{test}} \right]} \equiv \text{E}_{\mathcal{D}/\mathcal{R}}\left[\left(E_{(x,g) \sim P_{\text{test}} }\left[ \delta_{g,\;\widehat{G}^\mathcal{D}(x)}\right] - \overline{\text{acc}}\left[\widehat{G}, P_{\text{test}} \right] \right)^2\right],
\]
can be estimated in the same spirit by
\[
\widehat{\sigma}^2_{\text{acc}\left[\widehat{G}, P_{\text{samp}} \right]} = \frac{1}{n_{\text{splits}}-1} \sum_{i=1}^{n_{\text{splits}}} \left(\widehat{\text{acc}}\left[\widehat{G}^{\mathcal{D}_i},  P_{\text{samp}} \right]  -  \widehat{\overline{\text{acc}}}\left[\widehat{G}, P_{\text{samp}} \right]  \right)^2.
\]

Along with these metrics we can study the confusions between two classes using normalized confusion matrix, $\widehat{\overline{m}}$, that measures the frequency of predicting $\widehat{G}^\mathcal{D}(x)=j$ given that the real class is $G=i$, i.e.
\begin{align}
\widehat{\overline{m}}_{ij} \left[\widehat{G},  P_{\text{samp}} \right] = \frac{1}{n_{\text{splits}}} \sum_{i=1}^{n_{\text{splits}}} \sum_{(x,v,g) \in \mathcal{D}^{\text{t-c}}_i }  \frac{1}{n^{(\text{photos})}_{g v}} \cdot \frac{1}{n^{(\text{pots})}_g }\cdot  \delta_{g,i} \cdot \delta_{\widehat{G}^{\mathcal{D}_i}(x),j}. \label{confusionmatrix}
\end{align} 
Notice that $\sum_j  \widehat{\overline{m}}_{ij} = 1$.


\section{Results}
\label{results}

In this section, we show the results of the metrics described in Section \ref{metrics} for the four different architectures' ({\it Inception v3}, {\it Resnet50 v2},  {\it Mobilenet v2} and {\it VGG19}) and four different pretrainings ({\it ImageNet, matplotlib, blender1} and {\it blender2}). We demonstrate the positive effect of the different pretrainings with simulated photographs in all architectures performance. The effect of pot damage and the photo viewpoint are also assessed as they are important factors to be considered.

\subsection{Accuracy}
\label{accuracy}

\begin{figure}[!tbp]
\centering
\begin{minipage}[b]{0.9\textwidth}
 \begin{minipage}[b]{0.499\textwidth}
    \includegraphics[width=\textwidth]{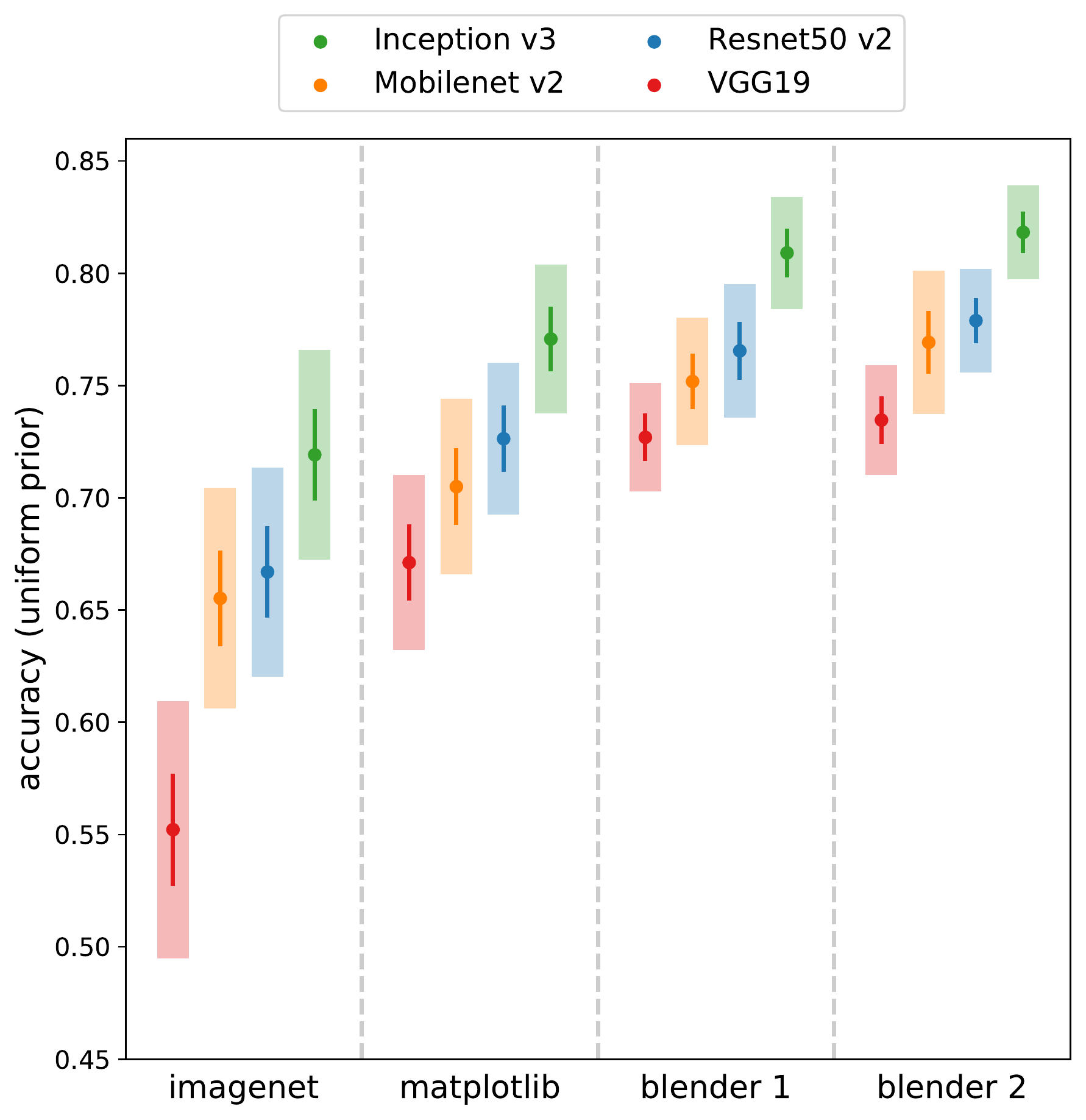}
  \end{minipage}
  \hfill
 \begin{minipage}[b]{0.499\textwidth}
    \includegraphics[width=\textwidth]{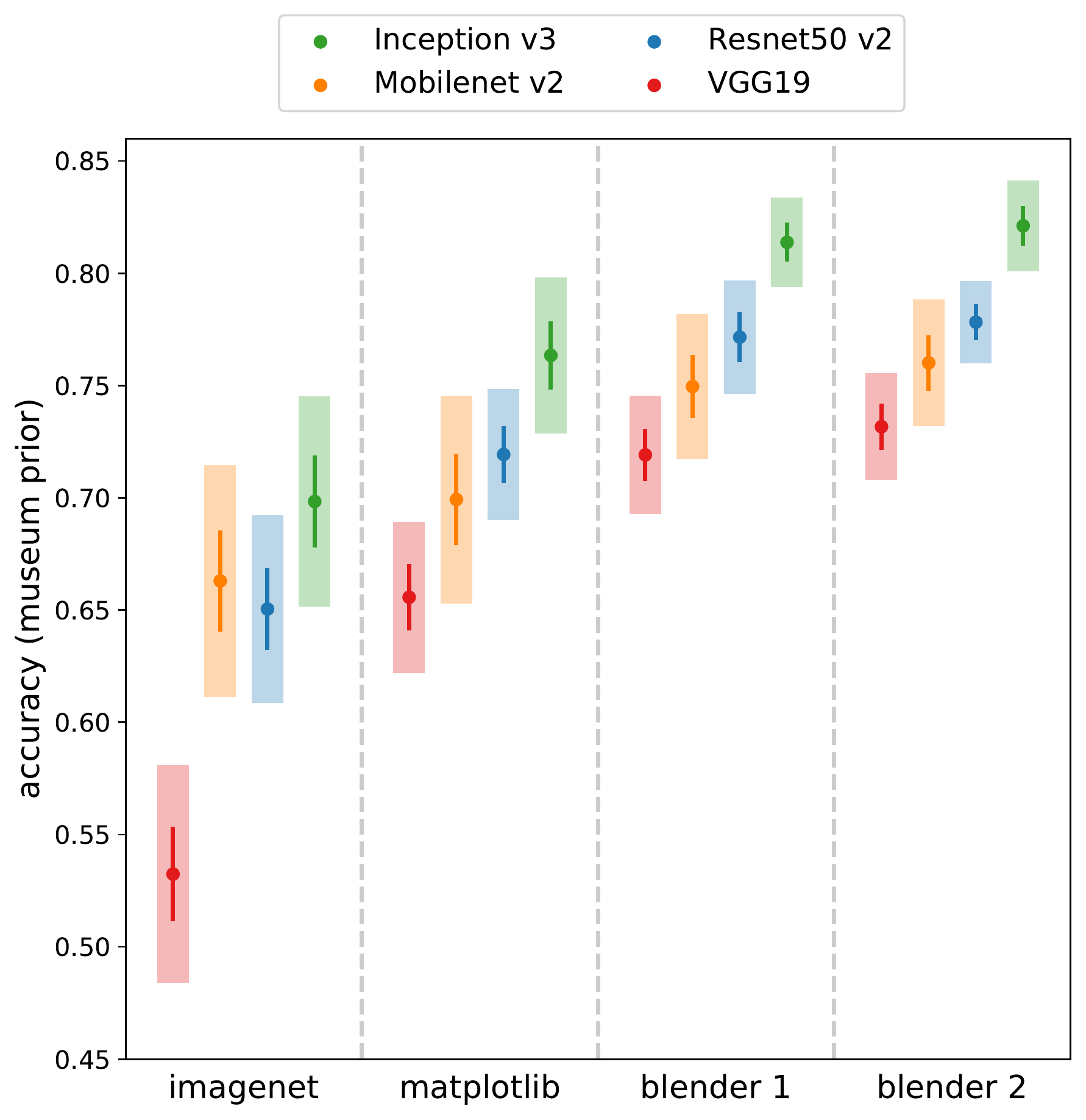}
  \end{minipage}
 \end{minipage}
\caption{\footnotesize{Two plots showing accuracy results for the four architectures considered:  {\it Inception v3} (green), {\it Resnet50 v2} (blue),  {\it Mobilenet v2} (orange) and {\it VGG19} (red). The figure on the left shows accuracy using the uniform prior, the one on the right using the MoL prior. The different pretraining regimes considered are displayed along the x-axis. Each point indicates the value of  $\widehat{\text{acc}}$ with its two sigma error band draw as a line. The transparent band shows  $\pm\hat{\sigma}_\text{acc}$.}}
\label{plot_acc_summary}
\end{figure}

Two tables summarizing the accuracy results for the 20 partitions can be found in Table \ref{unbiased_acc_table}, for the uniform prior, and in Table \ref{biased_acc_table}, for the MoL prior. The best results are obtained by the architecture with {\it Inception V3} backbone that surpasses the $80\%$ accuracy with the best pretraining (blender2). As it can be seen, the pretraining with simulated photographs has a considerable positive effect irrespective of architectures. A graphical visualization of these results can be seen in Figure \ref{plot_acc_summary}.

\begin{figure}[!tbp]
\centering
\begin{minipage}[b]{\textwidth}
 \begin{minipage}[b]{0.499\textwidth}
    \includegraphics[width=\textwidth]{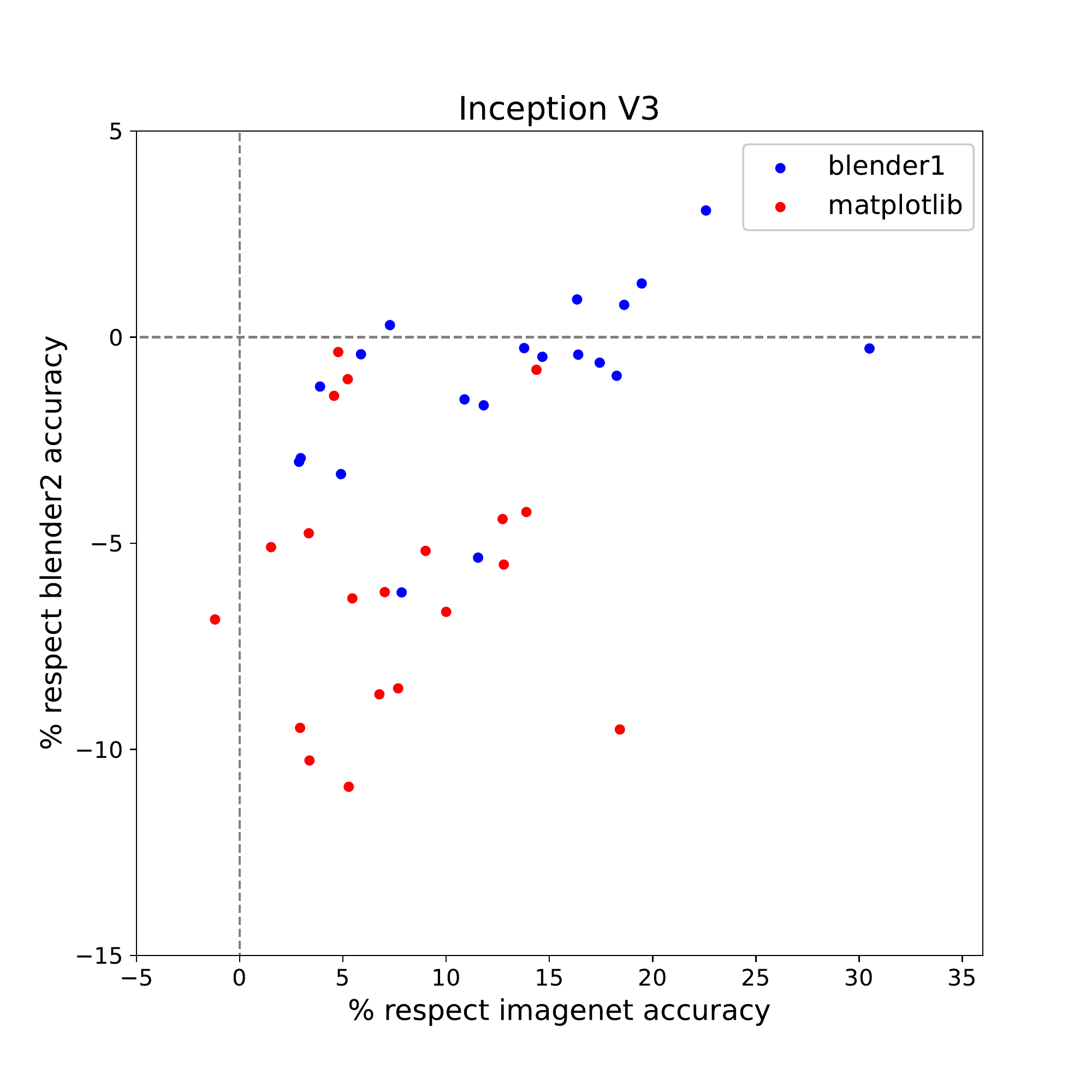}
  \end{minipage}
  \hfill
 \begin{minipage}[b]{0.499\textwidth}
    \includegraphics[width=\textwidth]{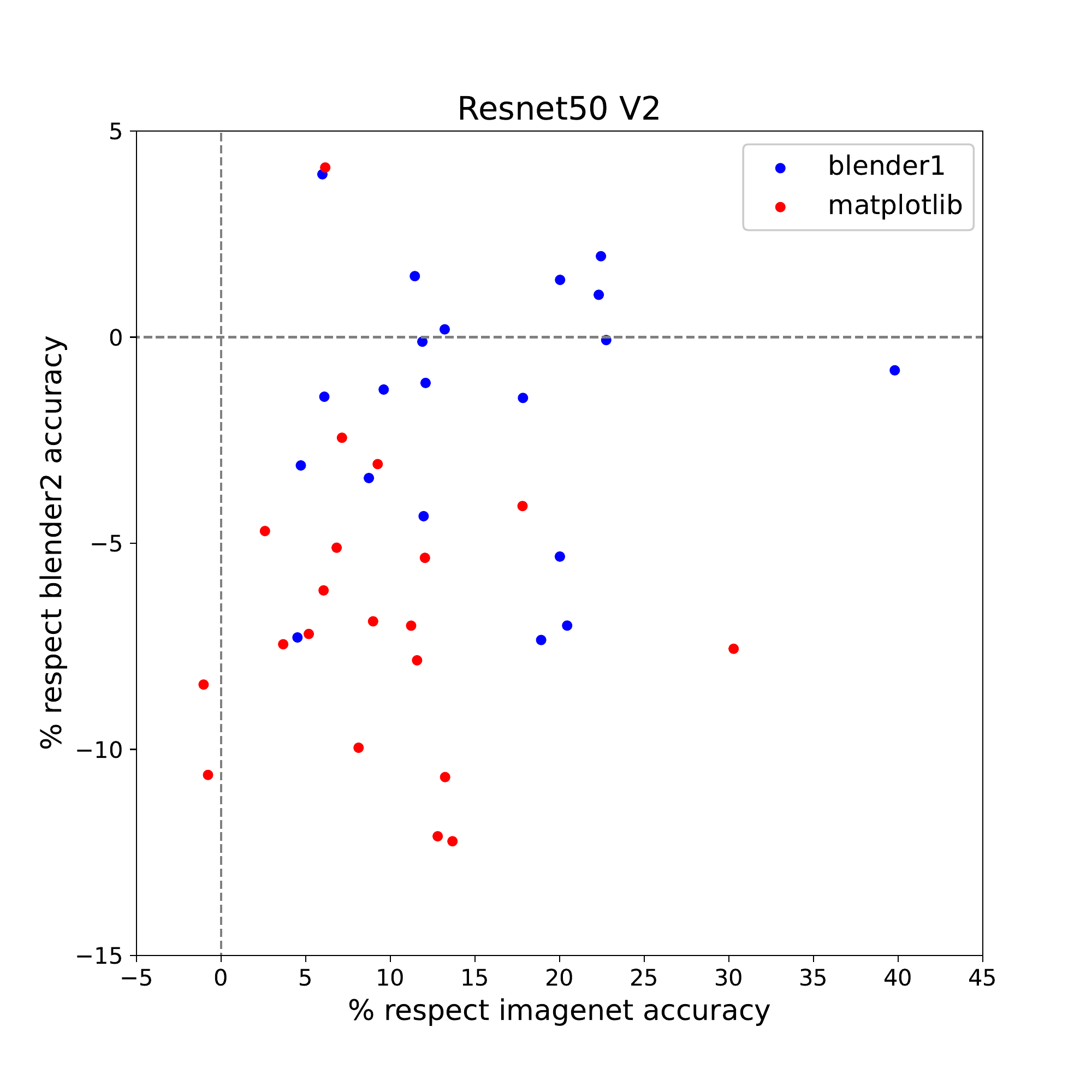}
  \end{minipage}
\end{minipage}
\begin{minipage}[b]{\textwidth}
 \begin{minipage}[b]{0.499\textwidth}
    \includegraphics[width=\textwidth]{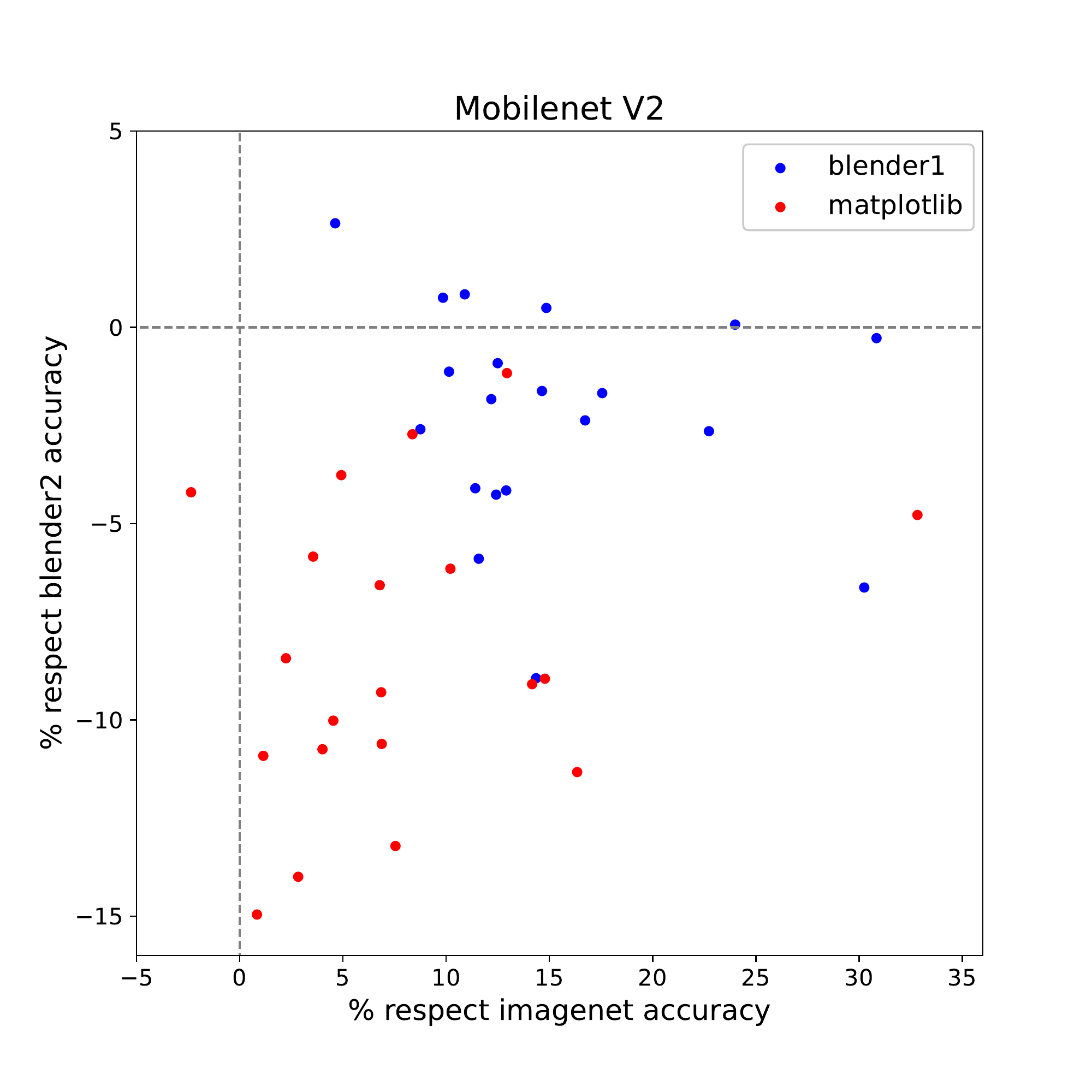}
  \end{minipage}
  \hfill
 \begin{minipage}[b]{0.499\textwidth}
    \includegraphics[width=\textwidth]{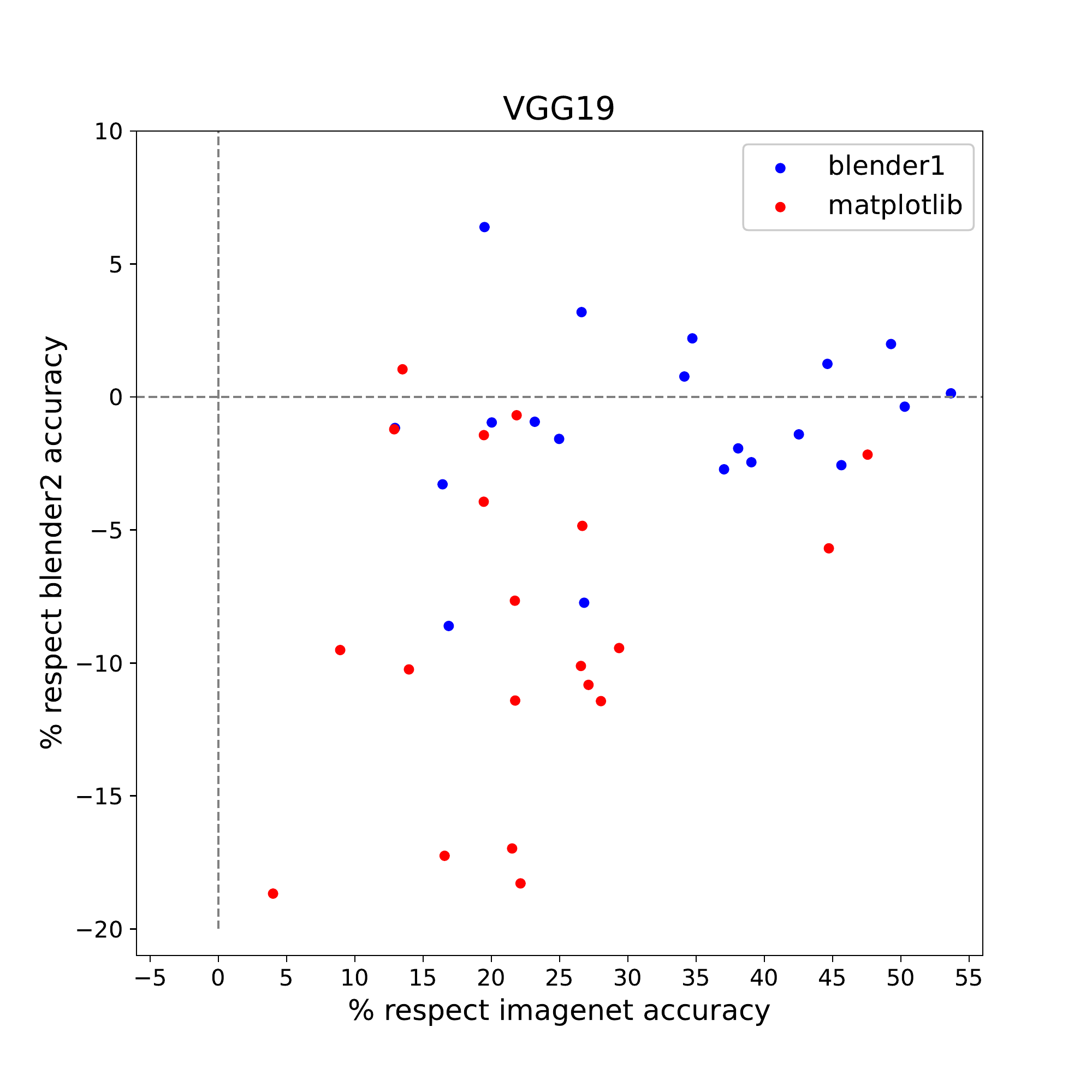}
  \end{minipage}
\end{minipage}
\caption{\footnotesize{ Four plots showing the accuracy results of the pretraining regimes relative to each other, using the uniform prior, for the four networks:  {\it Inception v3} (superior left), {\it Resnet50 v2} (superior right),  {\it Mobilenet v2} (inferior left) and {\it VGG19} (inferior right). Each point represents a single training-validation-test partition with the system pretrained with the blander dataset (blue) and Matplotlib dataset (red) where their positions are given as the relative percentage with respect to the uniform prior accuracy in the same partition for blender2 pretraining (Y axis) and imagenet pretraining (X axis).  As can be seen, most of the points are located above $0\%$  with respect to imagenet accuracy and below for blender2.}}
\label{unbiased_folds}

\end{figure}

The best model for each architecture is the one pretrained with blender2 dataset. The average accuracy is within the two sigma region of  the models pretrained with blender1, though systematically higher for all architectures. The estimated variance, $\hat{\sigma}_{\text{acc}}$, for blender2 pretraining is slightly smaller as well. A comparison of the results for each partition can be found in Figure \ref{unbiased_folds}, which also show that the pretraining with blender2 dataset could be slightly better than with blender1.

\begin{table}[!tbp]

\begin{minipage}[b]{0.9\textwidth}
\begin{center}
\bgroup
\def\arraystretch{1.5}
\begin{tabular}{|c|c|c|c|c|c|c|c|}
\hline
{\bf model} &   {\bf pretrain} &   {\bf $\widehat{\overline{\text{acc}}}$}  &    {\bf $\hat{\sigma}_\text{acc}$} &   {\bf $\text{min}_{\widehat{\text{acc}}}$} &    {\bf  $\text{max}_{\widehat{\text{acc}}}$} \\
\hline\hline
Inception v3 &    blender2 & 0.8182  $\pm$   0.0089 & 0.021 & 0.78 &  0.87 \\
Inception v3  &    blender1 & 0.809  $\pm$         0.011 & 0.025 & 0.76 & 0.84 \\
Inception v3  &  matplotlib & 0.771  $\pm$         0.014 & 0.033 &   0.72 & 0.85 \\
Inception v3 &    imagenet & 0.719  $\pm$         0.020 & 0.047 &  0.61 & 0.82 \\
\hline
 Resnet50 v2 &    blender2 & 0.7789 $\pm$          0.0098 & 0.023 &0.74 &  0.82 \\
 Resnet50 v2 &    blender1 & 0.765 $\pm$          0.013 & 0.030 & 0.72 & 0.82 \\
 Resnet50 v2 &  matplotlib & 0.726 $\pm$        0.014 & 0.034 & 0.67 & 0.82 \\
 Resnet50 v2 &    imagenet & 0.667 $\pm$         0.020 & 0.047 & 0.55 & 0.77 \\
\hline
 Mobilenet v2 &    blender2 & 0.769 $\pm$          0.014 & 0.032 & 0.72 & 0.84 \\
 Mobilenet v2 &    blender1 & 0.752 $\pm$         0.012 & 0.028 & 0.70 & 0.82 \\
 Mobilenet v2 &  matplotlib &  0.705  $\pm$         0.017 & 0.039 & 0.65 & 0.78 \\
 Mobilenet v2 &    imagenet & 0.655  $\pm$         0.021 & 0.049 & 0.56 & 0.76 \\
\hline
 VGG19 &    blender2 & 0.735 $\pm$          0.010 & 0.025 &0.69 &  0.77 \\
 VGG19 &    blender1 & 0.727 $\pm$          0.010 & 0.024 & 0.69 & 0.78 \\
 VGG19 &  matplotlib & 0.671 $\pm$        0.017 & 0.039 & 0.60 & 0.74 \\
 VGG19 &    imagenet & 0.552$\pm$         0.024 & 0.057 & 0.47 & 0.65 \\
\hline
\end{tabular}

\egroup
\end{center}
\end{minipage}

\caption{\footnotesize{Summary of accuracy results using the uniform prior. The different statistics are computed over the 20 test  results for each partition: $\overline{\text{acc}}$ is the average accuracy with its two sigma bootstrap error, $\sigma_\text{acc}$ is the variance estimation, $\text{min}_\text{acc}$ is the minimum result and $\text{max}_\text{acc}$ is the maximum. Each statistic is computed for the four architectures ({\it model}) and all the different initial weights considered ({\it pretrain}).  }}
\label{unbiased_acc_table}
\end{table}

\begin{table}[!tbp]
\begin{minipage}[b]{0.9\textwidth}
\begin{center}
\bgroup
\def\arraystretch{1.5}
\begin{tabular}{|c|c|c|c|c|c|c|c|}
\hline
{\bf model} &   {\bf pretrain} &   {\bf $\widehat{\overline{\text{acc}}}$}  &    {\bf $\hat{\sigma}_\text{acc}$} &   {\bf $\text{min}_{\widehat{\text{acc}}}$} &    {\bf  $\text{max}_{\widehat{\text{acc}}}$} \\
\hline\hline
Inception v3 &    blender2 & 0.8211 $\pm$         0.0087 & 0.020 & 0.78 & 0.86 \\
Inception v3 &    blender1 & 0.8138 $\pm$        0.0085 &  0.020 & 0.78 & 0.85 \\
Inception v3 &  matplotlib & 0.763 $\pm$           0.015 & 0.035 & 0.71 & 0.82 \\
Inception v3 &    imagenet & 0.698 $\pm$        0.020 & 0.047 & 0.62 & 0.80\\

\hdashline
 Resnet50 v2 &    blender2 & 0.7782 $\pm$          0.0078 & 0.018 & 0.74 & 0.81 \\
 Resnet50 v2  &    blender1 & 0.772 $\pm$          0.011 &  0.025 & 0.73 & 0.81 \\
 Resnet50 v2  &  matplotlib & 0.719 $\pm$         0.012 & 0.029 & 0.66 & 0.79 \\
 Resnet50 v2  &    imagenet &  0.651 $\pm$          0.018 & 0.042 &  0.54 & 0.74 \\

\hdashline
Mobilenet v2 &    blender2 & 0.760 $\pm$          0.012 & 0.028 & 0.71 & 0.81 \\
Mobilenet v2 &    blender1 & 0.750 $\pm$          0.014 &  0.032 & 0.69 & 0.81 \\
Mobilenet v2 &  matplotlib & 0.699 $\pm$         0.020 & 0.046 & 0.59 & 0.77 \\
Mobilenet v2 &    imagenet &  0.663 $\pm$          0.022 & 0.052 &  0.55 & 0.74 \\

\hdashline
 VGG19  &    blender2 & 0.732 $\pm$          0.010 & 0.024 & 0.68 & 0.77 \\
 VGG19  &    blender1 & 0.719 $\pm$          0.011 &  0.026 & 0.67 & 0.77 \\
 VGG19  &  matplotlib & 0.656 $\pm$         0.014 & 0.034 & 0.58 & 0.72 \\
 VGG19  &    imagenet &  0.532 $\pm$          0.021 & 0.048 &  0.46 & 0.61 \\

\hline
\end{tabular}
\egroup
\end{center}
\end{minipage}
\caption{\footnotesize{Summary of accuracy results using the MoL prior. The different statistics are computed over the  20 test  results for each partition: $\overline{\text{acc}}$ is the average accuracy with its two sigma bootstrap error, $\sigma_\text{acc}$ is the variance estimation, $\text{min}_\text{acc}$ is the minimum result and $\text{max}_\text{acc}$ is the maximum. Each statistic is computed for the four architectures ({\it model}) and all the different initial weights considered ({\it pretrain}).  }}
\label{biased_acc_table}
\end{table}

\subsection{Confusion matrix}

The normalized confusion matrix defined in equation (\ref{confusionmatrix}) shows the main instances of error between two classes of pot. To simplify discussion of the results we will focus on the best architecture, {\it Inception v3}, for the extreme cases of ImageNet and blender2 pretrainings. The normalized confusion matrix, $\widehat{\overline{m}}_{ij}$, can be seen in Table \ref{norm_cm_imagenet} and Table \ref{norm_cm_blender2} for each of these, respectively. We will consider off-diagonal elements a major instance of confusion when they exceed double the expected probability if the incorrect identifications would have been uniformly distributed across the classes, i.e.:
\begin{align}
\widehat{\overline{m}}_{ij} \left[\widehat{G},  P_{\text{samp}} \right] > 2 \; \frac{1-\widehat{\overline{m}}_{ii} \left[\widehat{G},  P_{\text{samp}} \right]}{n_{\text{classes}}-1},
\label{major_confusion}
\end{align}
with $n_{\text{classes}} = 9$ the number of classes in our dataset.

We can see that pretraining with the simulation photographs seems to have resolved some confusions like {\it Dr35} with {\it Dr27} or {\it Dr18} with {\it Dr36}. However, there are other mix-ups that seem to be difficult to correct, for example {\it Dr35} with {\it Dr36}. This latter confusion was to be expected as these vessel forms have near identical shape and decoration, the main difference being size and height-radius ratio \cite{webster1996roman}. A similar thing happens with the confusion between the two decorated forms {\it Dr37} and {\it Dr29}. Other confusions are more difficult to explain, e.g. {\it Dr38} with {\it Dr36}.

\begin{table}[!tbp]
\centering
\begin{minipage}[b]{0.9\textwidth}
\centering
\resizebox{\textwidth}{!}{
\bgroup
\def\arraystretch{1.5}
\begin{tabular}{|c|c|c|c|c|c|c|c|c|c|}
\hhline{~---------}
\multicolumn{1}{c|}{} & \multicolumn{9}{c|}{{\bf Predicted class} }\\
\hline
{\bf Real class} &  {\bf Dr18} &  {\bf Dr24-25} &   {\bf Dr27} &   {\bf Dr29} &   {\bf Dr33} &  {\bf  Dr35} &   {\bf  Dr36} &   {\bf  Dr37} &   {\bf  Dr38} \\
\hline
    {\bf Dr18}& \cellcolor{green!25}  63 &        4 &     2 &     2 &     6 &     2 &   \cellcolor{red!25} 14 &     3 &     3 \\
\hline
 {\bf Dr24-25} &     5 &      \cellcolor{green!25}  76 &     5 &     1 &     3 &     3 &     2 &     1 &     4 \\
\hline
   {\bf  Dr27} &     2 &        4 &    \cellcolor{green!25} 74 &     2 &     2 &    \cellcolor{red!25} 9 &     2 &     2 &     4 \\
\hline
   {\bf  Dr29}&     2 &        2 &     3 &    \cellcolor{green!25} 74 &     2 &     1 &     4 &     \cellcolor{red!25}7 &     5 \\
\hline
   {\bf  Dr33}&     \cellcolor{red!25} 3 &        2 &     2 &     1 &    \cellcolor{green!25} 89 &     2 &     0 &     1 &     1 \\
\hline
  {\bf   Dr35}&     1 &        4 &   \cellcolor{red!25} 11 &     1 &     1 &   \cellcolor{green!25}  67 &    \cellcolor{red!25} 11 &     1 &     3 \\
\hline
    {\bf Dr36}  &     8 &        4 &     2 &     4 &     1 &   \cellcolor{red!25} 12 &    \cellcolor{green!25} 60 &     2 &     8 \\
\hline
   {\bf  Dr37} &     2 &     \cellcolor{red!25}   9 &     1 &   \cellcolor{red!25} 11 &     3 &     1 &     1 &   \cellcolor{green!25}  68 &     4 \\
\hline
  {\bf   Dr38} &     1 &        6 &     2 &     3 &     2 &     2 &   \cellcolor{red!25}  7 &     1 &   \cellcolor{green!25}  76 \\
\hline
\end{tabular}
\egroup}

\end{minipage}

\caption{\footnotesize{Normalized confusion matrix of equation (\ref{confusionmatrix}) for {\it Inception v3} pretrained with the Imagenet dataset. Diagonal elements are highlighted in green, whereas major instances of confusion are marked in red. Major instances of confusion are those off-diagonal elements which exceed double the expected probability if the incorrect identifications would have been uniformly distributed across the classes, see equation (\ref{major_confusion}). }}
\label{norm_cm_imagenet}

\vspace{2em}

\begin{minipage}[b]{0.9\textwidth}

\centering
\resizebox{\textwidth}{!}{
\bgroup
\def\arraystretch{1.5}
\begin{tabular}{|c|c|c|c|c|c|c|c|c|c|}
\hhline{~---------}
\multicolumn{1}{c|}{} & \multicolumn{9}{c|}{{\bf Predicted class} }\\
\hline
{\bf Real class} &  {\bf Dr18} &  {\bf Dr24-25} &   {\bf Dr27} &   {\bf Dr29} &   {\bf Dr33} &  {\bf  Dr35} &   {\bf  Dr36} &   {\bf  Dr37} &   {\bf  Dr38} \\
\hline
    {\bf Dr18} &  \cellcolor{green!25} 81 &        2 &     1 &     3 &     2 &     1 &     \cellcolor{red!25} 7 &     2 &     1 \\
\hline
 {\bf Dr24-25}&   3 &      \cellcolor{green!25}  80 &     1 &     2 &     2 &     3 &     2 &     2 &   \cellcolor{red!25}  6 \\
\hline
   {\bf  Dr27} & 1 &        2 &    \cellcolor{green!25} 87 &     1 &   3 &    3 &     1 &     1 &     1 \\
\hline
   {\bf  Dr29}&     2 &        2 &     1 &  \cellcolor{green!25}  81 &     2 &     1 &     2 &    \cellcolor{red!25} 7 &     3 \\
\hline
   {\bf  Dr33} &     0 &        1 &     0 &     1 &  \cellcolor{green!25}  95 &     1 &     0 &     0 &     0 \\
\hline
  {\bf   Dr35}&     1 &        0 &     4 &     1 &     1 &  \cellcolor{green!25}   80 &   \cellcolor{red!25} 10 &     0 &     2 \\
\hline
    {\bf Dr36} &     6 &        2 &     0 &     1 &     1 &   \cellcolor{red!25} 11 &  \cellcolor{green!25}  72 &     1 &    6 \\
\hline
   {\bf  Dr37}  &     1 &       \cellcolor{red!25} 6 &     1 &     \cellcolor{red!25}7 &     1 &     0 &     0 &  \cellcolor{green!25}  82 &     3 \\
\hline
  {\bf   Dr38} &     2 &        3 &     2 &     1 &     1 &     3 &     \cellcolor{red!25}8 &     1 &   \cellcolor{green!25}  79 \\
\hline
\end{tabular}

\egroup}

\end{minipage}

\caption{\footnotesize{Normalized confusion matrix of equation (\ref{confusionmatrix}) for {\it Inception v3} pretrained with the blender2 dataset. Diagonal elements are highlighted in green, whereas major instances of confusion are marked in red. Major instances of confusion are those off-diagonal elements which exceed double the expected probability if the incorrect identifications would have been uniformly distributed across the classes, see equation (\ref{major_confusion}). }}
\label{norm_cm_blender2}
\end{table}

\subsection{Effects of damage}
\label{damage_section}
As we have previously discussed, some of the pots are severely damaged. Even if a complete profile can be recovered from the vessel, extreme damage can adversely affect the classifier's predictions. In this section, we will evaluate the influence of damage on the performance.

Those vessels with large breaks affecting more than half of their rotational shapes were manually labelled as {\it damaged}. In order to evaluate the influence of damage, we measured the accuracy for {\it damaged} and {\it non damaged} vessels at each of the 20 train-validation-test partitions. The number of {\it damaged} pots in this dataset is small, however, and can vary at each partition. It is even possible not to have any samples for some classes at a given partition. Therefore, we weight the accuracy for each class, proportional to the number of {\it damaged} vessels. Thus, with $\widehat{\text{acc}}^{(\text{dam})}_i$ and $\widehat{\text{acc}}^{(\text{nodam})}_i$ being the accuracy described in  (\ref{acc_single}) restricted to {\it damaged} or {\it non damaged}) pots of class $i$. The accuracy per pot class can be seen in Table \ref{inception_damaged} ({\it damaged}) and in Table \ref{inception_non_damaged} ( {\it non damaged}) for the best model, i.e. the model based on {\it Inception v3} architecture. As we can see there is a huge general fall in accuracy for {\it damaged} pots, which pretraining seems to help to mitigate, at least in the case of the {\it Inception v3} model.

\begin{table}[!tbp]

\begin{minipage}[b]{0.9\textwidth}
\centering
\resizebox{0.8\textwidth}{!}{
\bgroup
\def\arraystretch{1.5}
\begin{tabular}{|c|c|c|c|c|c|}
\hhline{~-----}
\multicolumn{1}{c|}{} & \multicolumn{5}{c|}{Damaged} \\
\hline
   class &  ImageNet &  matplotlib &  blender1 &  blender2 &  $\overline{n^\text{dam}}$  \\
\hline\hline
    Dr18 &      0.48 &        0.52 &      0.55 &       0.60 &         13.60  \\
 Dr24-25 &      0.31 &        0.56 &      0.41 &      0.36 &        0.20 \\
    Dr27 &      0.53 &         0.60 &      0.69 &      0.69 &        4.75 \\
    Dr29 &      0.69 &        0.73 &      0.77 &      0.72 &        5.55  \\
    Dr33 &      0.81 &        0.85 &      0.91 &      0.93 &        1.95  \\
    Dr35 &        -  &      -      &    -      &    -&             0.00    \\
    Dr36 &      0.53 &        0.62 &       0.60 &      0.63 &          3.05 \\
    Dr37 &      0.56 &        0.33 &      0.69 &      0.76 &        0.40  \\
    Dr38 &      0.79 &        0.74 &      0.78 &      0.78 &        1.05 \\

\hline
\end{tabular}
\egroup}
\end{minipage}
\caption{\footnotesize{Table showing accuracy results of the {\it Inception v3} model for vessels labelled {\it damaged}. $\overline{n^\text{dam}}$   indicates the average number of damaged pots in each class at a split in the test set, thus lower numbers are less reliable. Notice that the results for damaged vessels of class {\it Dr35} could not be computed as there are no damaged samples.}}
\label{inception_damaged}

\end{table}

\begin{table}[!tbp]
\begin{minipage}[b]{0.9\textwidth}
\centering
\resizebox{0.8\textwidth}{!}{
\bgroup
\def\arraystretch{1.5}
\begin{tabular}{|c|c|c|c|c|c|}
\hhline{~-----}
\multicolumn{1}{c|}{} & \multicolumn{5}{c|}{Non-Damaged}\\
\hline
   class & ImageNet &  matplotlib &  blender1 &  blender2 &  $\overline{n^\text{no dam}}$ \\
\hline\hline
    Dr18 &     0.70 &        0.81 &       0.90 &       0.90 &         31.40 \\
 Dr24-25 &          0.86 &        0.89 &      0.85 &      0.89 &        1.80 \\
    Dr27 &            0.8 &         0.90 &      0.92 &      0.92 &         17.25 \\
    Dr29 &           0.79 &        0.82 &      0.89 &      0.88 &        6.45 \\
    Dr33 &           0.92 &        0.91 &      0.95 &      0.96 &        5.05 \\
    Dr35 &           0.67  &    0.69   &    0.78  &     0.80  &            6.00      \\
    Dr36 &           0.64 &        0.72 &      0.79 &      0.77 &          5.95 \\
    Dr37 &          0.64 &        0.86 &      0.79 &      0.82 &        2.60 \\
    Dr38 &        0.75 &        0.81 &      0.77 &      0.82 &       0.95\\

\hline
\end{tabular}
\egroup}
\end{minipage}
\caption{\footnotesize{Table showing accuracy results of the {\it Inception v3} model for {\it non-damaged} vessels.  $\overline{n^\text{no dam}}$   indicates the average number of {\it non-damaged} pots in each class at a split in the test set, thus lower numbers are less reliable.}}
\label{inception_non_damaged}

\end{table}

\subsection{Effects of viewpoint}
\label{view_section}

Another interesting division of the dataset is the viewpoint from which the photograph was taken. We have manually labelled three viewpoints, namely: {\it standard} view, {\it zenith} view, {\it flipped} vessel. An example of each viewpoint can be seen in Figure \ref{types_photos}. In both {\it standard} and {\it zenith} views the vessel is standing on its base, whereas in {\it flipped} view it is supported by its rim.

\begin{figure}[!tbp]
\centering
 \begin{minipage}[b]{0.7\textwidth}
    \includegraphics[width=\textwidth]{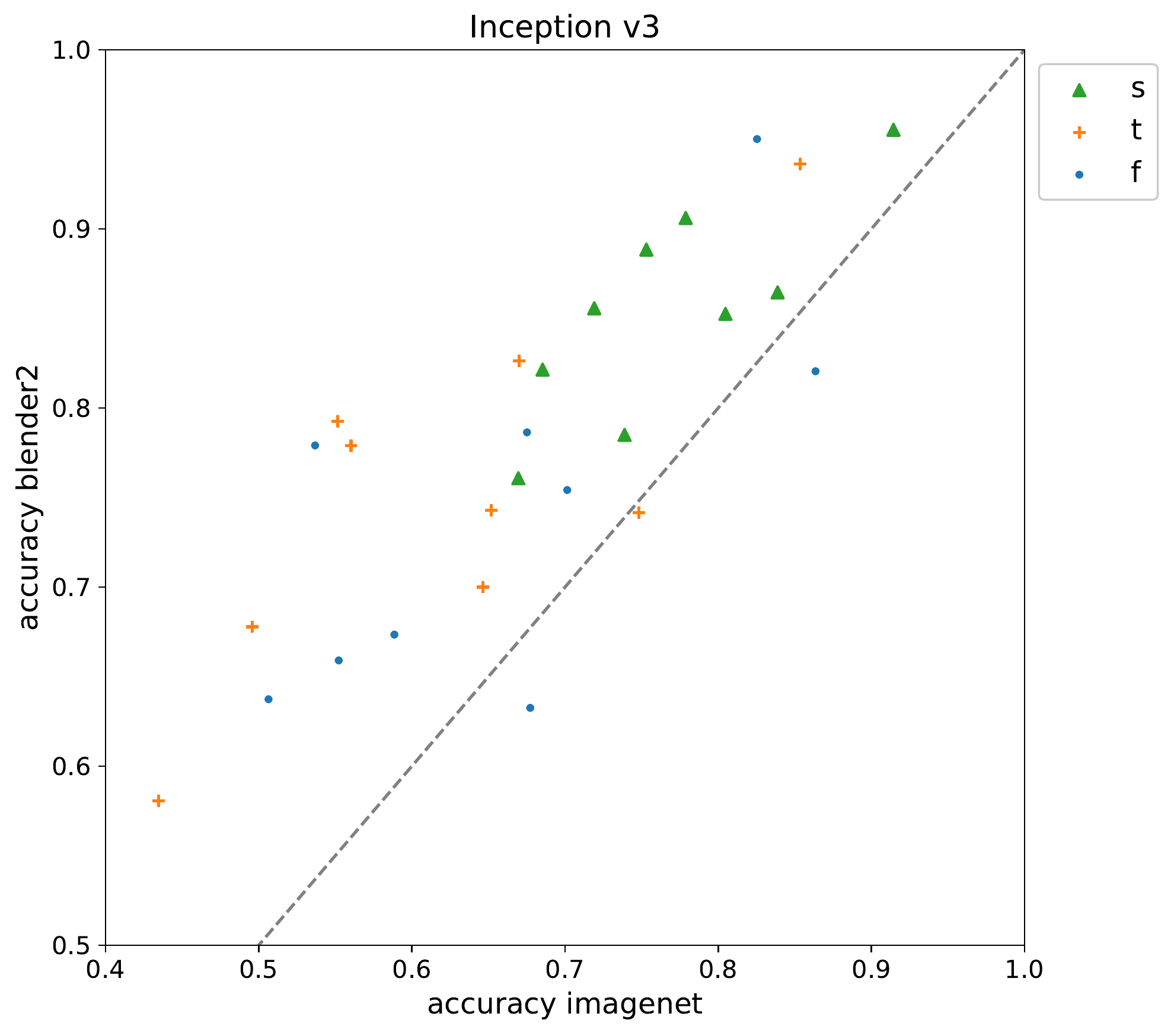}
  \end{minipage}
  \caption{\footnotesize{Plot detailing the accuracy of the {\it Inception v3} model across the different viewpoints and comparing performance with blender2 and Imagenet pretraining.  Each point represents the accuracy for each pot class and viewpoint: {\it standard} (green triangle), {\it zenith} (orange cross), {\it flipped} (blue point). The dashed grey line shows the point where the accuracies of both pretraining regimes are the same. The fact that most of the points are well above these lines shows how pretraining with blender2 improves the performance in general.  Notice also that the {\it standard} view shows generally a better performance in both models in comparison with the {\it zenith} and {\it flipped} views.} }
\label{view_blender2_imagenet}
\end{figure}

\begin{figure}[!tbp]
\centering
 \begin{minipage}[b]{0.8\textwidth}
    \includegraphics[width=\textwidth]{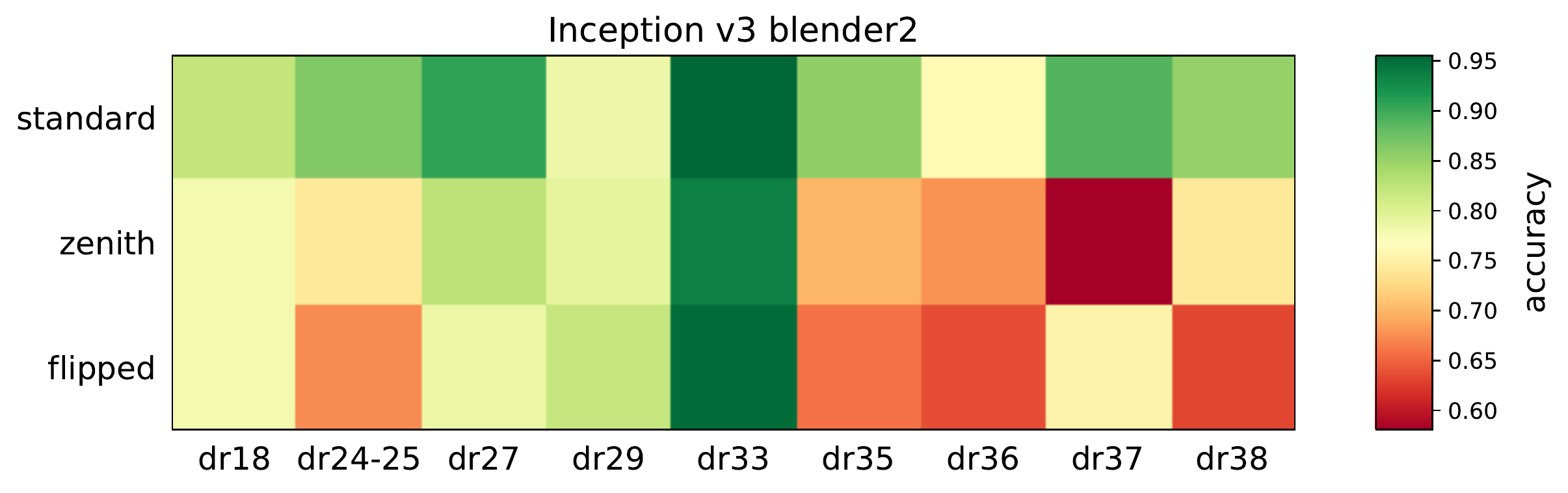}
  \end{minipage}
  \caption{\footnotesize{Plot showing the variation in the accuracy of the {\it Inception v3} model pretrained with blender2 across viewpoints and different pot classes.}}
\label{view_class}
\end{figure}

A comparison between the difference in performance per viewpoint of {\it Inception v3} with blender2 and {\it ImageNet} pretrainings is shown in Figure \ref{view_blender2_imagenet}. Improvement in accuracy seems to have a similar intensity for all viewpoints. This global shift seems to be constant for the whole {\it ImageNet} accuracy range. Finally, a comparison of the accuracy per pot class and view for the best model {\it Inception v3} with blender2 pretraining can be found in Figure \ref{view_class}.

The {\it zenith} view is a special viewpoint as the profile is very hard to see. This fact together with the lower occurrence might explain its worse performance compared to other viewpoints. On the other hand, although the {\it flipped} view is a very nice viewpoint to get an idea of the profile, we still have a score that is worse than the {\it standard} view. A possible explanation for this result is the imbalance in the number of images. Across all photographs taken at the MoL, including the ones used for this experiment, the {\it standard} view outnumbers the {\it flipped} photographs six to one. Thus, the features created by the machines might be more likely to focus on the {\it standard} images.

\subsection{Reality gap}
\label{reality_gap}

An interesting question that arises from the different pretrainings is whether the neural nets trained with only simulated images are good enough models for classifying real photographs. The difference between the performance of a model trained with real images and the same model trained with simulations is usually called the {\it reality gap} \cite{tobin2017domain, collins2020traversing}.

\begin{figure}[t]
\centering
\begin{minipage}[b]{0.7\textwidth}
\centering
\includegraphics[width=\textwidth]{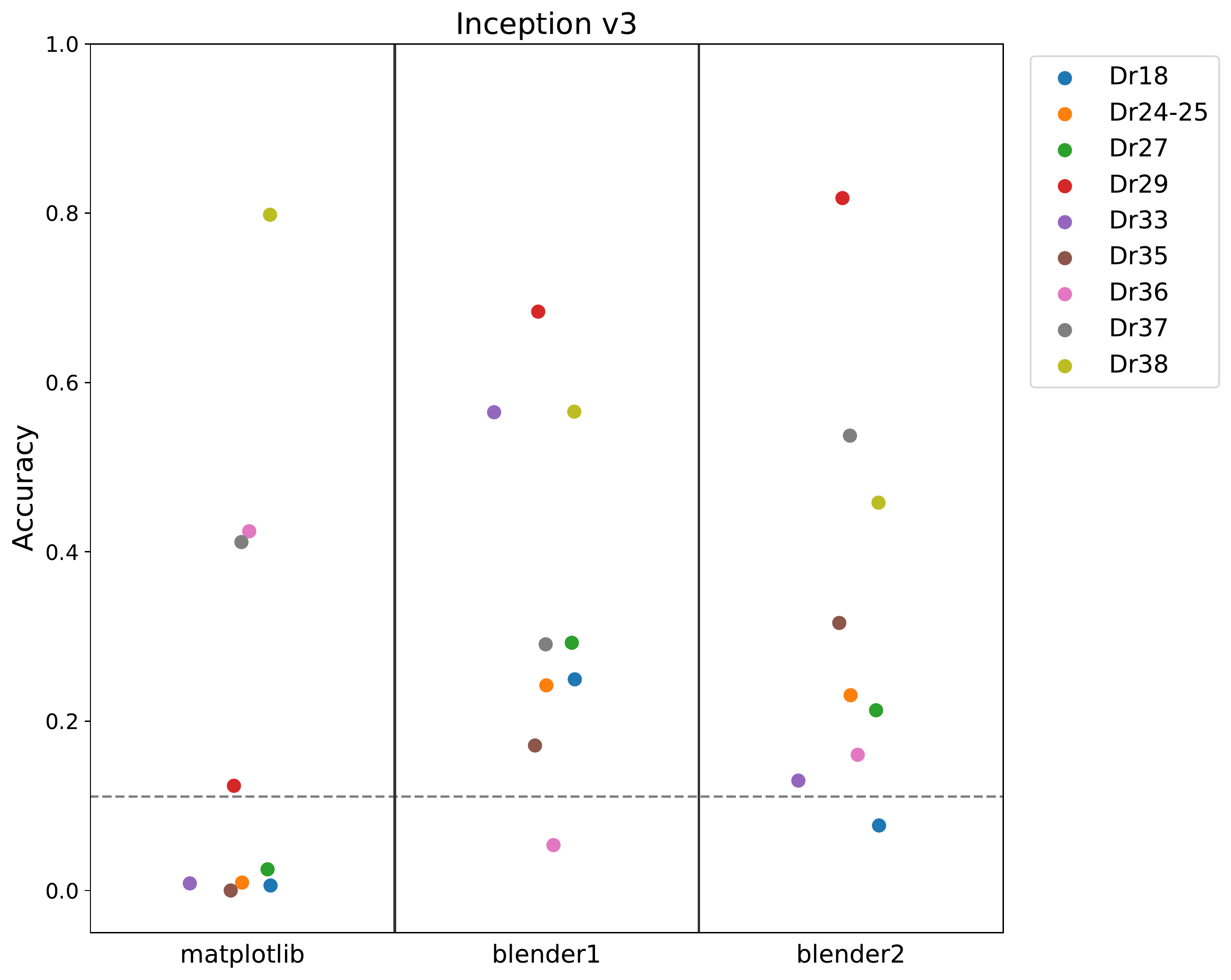}
 \end{minipage}
\caption{\footnotesize{ Plot of the {\it Inception v3} model accuracy for each pretraining for the different pot classes. The horizontal dashed grey line points out the random assignment accuracy $1/9$.}}
\label{reality_gap_img}
\end{figure}

In Figure \ref{reality_gap_img} comparisons among the accuracies for the different pot classes are shown for each simulated dataset and architecture. As we can see there are huge differences in the performance of Matplotlib and Blender datasets. This is to be expected as the quality of the images using Blender was improved substantially, see Figure \ref{classes_imgs}. It is also interesting to notice the huge variability per pot class and, how some profiles always have a good performance such as {\it Dr38}, while others like {\it Dr18} usually have poor results, though the latter is perhaps not very surprising as this concerns a class into which multiple types have been amalgamated, potentially making it more difficult for the classifier to extract distinguishing features. An aggregated comparison can be found in Table \ref{reality_gap_summary} for {\it Inception v3} model, where we can see that Blender datasets outperform Matplotlib.

\begin{table}[!tbp]
\centering
\begin{minipage}[b]{0.7\textwidth}
\centering
\resizebox{\textwidth}{!}{
\bgroup
\def\arraystretch{1.5}
\begin{tabular}{|c|c|c|c|c|c|c|}
\hhline{~------}
\multicolumn{1}{c|}{} & \multicolumn{2}{c|}{Matplotlib} &  \multicolumn{2}{c|}{blender1} & \multicolumn{2}{c|}{blender2}\\
\hline
   model &  mean&  std  &  mean &  std  &mean  &  std  \\
\hline\hline
  Inception v3&  0.20  & 0.28 &     0.35 &   0.21 &     0.33 &   0.24 \\
\hline
\end{tabular}
\egroup
}
\end{minipage}

\caption{\footnotesize{Table showing summary statistics for the performance of the {\it Inception v3 } based model trained only with the simulated images of datasets: matplotlib, blender1, blender2. The mean corresponds to an estimation of the uniform prior accuracy defined in Section \ref{accuracy}. We can see the improvement in blender datasets compared with matplotlib dataset.}}
\label{reality_gap_summary}
\end{table}


\section{Conclusions}\label{Conclusions}

In this article we have proposed that creating synthetic datasets can be used as a way of augmenting inherently limited real-world datasets. Through this dataset creation expert knowledge beyond the identification of real artefacts can be incorporated in the process of machine learning from scarce information. Furthermore, we have demonstrated, though experimentation with different network architectures and consideration of the impact of photograph viewpoint and damage to the original vessels, that a hybrid approach using synthetic data to pre-train a machine that is subsequently trained with real-world data has the potential to transcend the possibilities of a machine trained exclusively on real-world data. Since the number of fields where non-ideally sized or distributed datasets predominate is likely to significantly outnumber the fields where huge, balanced datasets are readily available, our results signal the expanded potential for machine learning applications in new fields.

Using a dataset of 5373 photos of 162 near-complete Roman {\it terra sigillata} vessels from the Museum of London, spread unequally over 9 typological classes, we have shown that it is possible to create a single image classifier that acceptably generalises beyond the training set. By digitizing pottery drawings, we were able to include expert knowledge in synthetic data generation, which allowed us not only to augment the size of the dataset, but also to somewhat counteract the skewed distribution in the real-world dataset. Both these aspects are crucial to the potential application of machine learning to domains where there are no perfect datasets. As compared to some other solutions to the problem of classifying archaeological ceramics, such as the ArchAIDE project \cite{itkin2019computational}, \cite{anichini2020developing}, \cite{anichiniautomatic}, we do not rely on the user’s input beyond the photograph. As such, we aim to include expert knowledge in training the classifier only, so that as little expertise as possible is required from any potential end user, making any eventual tool more widely usable.

We have also quantified how other factors such as the photograph viewpoint or excessive damage to the pot affect the results. The possible instances of confusion between classes were also discussed and show how simulation can help to reduce them in some cases. Due to the relatively uniform shapes and the relative simplicity of their simulations, vessel classification provides an excellent test bench for sim2real and 3D shape computer vision experiments. Here the limitations of the experiments described in this article need to be born in mind. The shape repertoire that the classifier was confronted with is rather limited and, even when augmented with synthetic images, the total dataset is rather small. It is therefore not the results of the performance that should be taken as an indication of the success of the model, nor the relative performance of the network architectures, but rather the improvement between different training regimes. It is these results that show the potential for using simulation to improve training image classifiers able to generalise beyond the training set in cases where only small or biased datasets are available, be this in archaeological ceramics or beyond.

\section*{Author contributions}

\textbf{Santos J. N\'u\~nez Jare\~no:} Conceptualization, Methodology, Software, Formal analysis, Investigation, Data Curation, Writing - Original Draft, Visualization. \textbf{Dani\"el P. van Helden:} Conceptualization, Methodology, Investigation, Data Curation, Writing Original Draft, Writing Review \& Editing, Project Administration. \textbf{Evgeny Mirkes:} Conceptualization, Validation, Writing Review\& Editing, Supervision. \textbf{Ivan Tyukin:} Conceptualization, Validation, Writing Review \& Editing, Supervision. \textbf{Penelope M. Allison:} Conceptualization, Writing Review \& Editing, Project Administration, Supervision, Funding Acquisition.

\section*{Funding}

The authors would like to thank the AHRC for funding the Arch-I-Scan project with grant AH/T001003/1.

\section*{Acknowledgements}

We would also like to thank the Museum of London and their staff (specifically Roy Stephenson and Nicola Fyfe) for granting us access to the collection, Fiona Seeley for expert pottery advice and identification, and Gabriel Florea and Alessandra Pegurri for helping with photography.

\section*{Conflicts of interests}

The authors declare no conflict of interest.

\bibliographystyle{plain}
\bibliography{bibliography}

\end{document}